\documentclass{article}

\PassOptionsToPackage{numbers}{natbib}

\usepackage[preprint]{neurips_2026}

\usepackage[utf8]{inputenc}
\usepackage[T1]{fontenc}
\usepackage{hyperref}
\hypersetup{linktoc=all}
\usepackage{url}
\usepackage{booktabs}
\usepackage{amsfonts}
\usepackage{amsmath}
\usepackage{nicefrac}
\usepackage{microtype}
\usepackage{xcolor}
\usepackage{multirow}
\usepackage{graphicx}
\usepackage[most]{tcolorbox}
\usepackage{enumitem}
\usepackage{tikz}
\usetikzlibrary{arrows.meta, positioning, calc}
\usepackage{float}
\usepackage{placeins}

\title{AI Achieves a Perfect LSAT Score}

\author{%
  Bonmu Ku\\
  \texttt{bku@alumni.harvard.edu}
}

\begin{document}

\maketitle
\addtocontents{toc}{\protect\setcounter{tocdepth}{-10}}

\begin{abstract}
This paper reports the first documented instance of a language model achieving a perfect score on an officially disclosed Law School Admission Test (LSAT). Controlled experiments on eight reasoning models show that varying the prompt, shuffling answer choices, and sampling multiple responses have no meaningful effect as drivers of performance. Ablating the thinking phase that models generate before answering, however, lowers frontier accuracy by up to 8 percentage points, predominantly in logical reasoning. Distilled models produce full thinking traces in the same format yet plateau far below frontier performance. A pilot process reward model fine-tuned via QLoRA on official LSAT explanations narrows this gap through Best-of-5 selection, with gains again predominantly in logical reasoning. The gatekeeper of elite legal education since 1948, the LSAT has not merely been passed but answered without a single error by models that reason. The upper bound of the cognitive capacities it has tested is no longer exclusive to human cognition.
\end{abstract}

\section{Introduction}
\subsection{The LSAT as a Benchmark for Logical Reasoning}

The Law School Admission Test (LSAT) stands as the most rigorous standardized examination of logical reasoning. The test requires no prior legal training. It instead requires exceptional reasoning capability to identify logical structures, evaluate arguments, and draw valid inferences from complex linguistic input. This exclusive measure of inferential skill makes the LSAT a particularly clean benchmark for evaluating the reasoning capabilities of language models as well.

Even the strongest human performers fall below the perfect score of 180, with a median of 174 among Harvard Law School admits. Language models have been measured against the same standard. GPT-3.5 scored 149 at the 40th percentile, and GPT-4 scored 163 at the 88th percentile \citep{openai2023gpt4}. Each generation of language models narrows the gap to the human ceiling. A perfect score, however, demands the capacity for extended multi-step reasoning and deliberation, a level of performance that has thus far remained unmet for language models.

\subsection{From Language Models to Reasoning Models}
\label{sec:background}

The leap from GPT-4 to the models that followed was not a continuation of prior scaling trends \citep{kaplan2020scaling, hoffmann2022training}. It marked the emergence of a fundamentally new inference paradigm: the reasoning model \citep{openai2024reasoning, deepseek2025r1, snell2024scaling}. The central capability was generating extended thinking traces at inference time before producing a final answer \citep{openai2024o1systemcard, kimi2025k15, gemini2025_25, anthropic2025claude37}. This advance developed incrementally, through three phases that moved explicit reasoning progressively deeper into the model.

\subsubsection*{Reasoning via Prompting}

Chain-of-thought prompting \citep{wei2022chain} dramatically improved performance on reasoning tasks by eliciting intermediate reasoning steps from the model. Zero-shot chain-of-thought \citep{kojima2022large} extended this result to the setting without few-shot exemplars, and self-consistency \citep{wang2023selfconsistency} further improved accuracy by majority-voting over multiple sampled reasoning paths. Still, the model did not yet internalize the reasoning process. Intermediate steps were elicited externally and remained sensitive to prompt phrasing.

\subsubsection*{Reasoning via Training}

STaR \citep{zelikman2022star} bootstrapped reasoning ability by generating chain-of-thought rationales, fine-tuning on those that yielded correct answers, and iterating. Quiet-STaR \citep{zelikman2024quietstar} generalized this by training models to produce internal rationales at every token position during general text prediction. On the evaluation side, outcome reward models (ORMs) \citep{cobbe2021training} scored candidate solutions by final-answer correctness, while process reward models (PRMs) \citep{lightman2023lets} scored each reasoning step independently and provided a stronger selection signal.

\subsubsection*{Reasoning via Inference}
Modern reasoning models are trained end-to-end via reinforcement learning to generate an internal chain of thought before producing a final answer \citep{openai2024reasoning, deepseek2025r1}. Inference-time compute scaling \citep{snell2024scaling} matched the gains of scaling model parameters by an order of magnitude when compute was optimally allocated. Group relative policy optimization (GRPO) \citep{shao2024deepseekmath}, originally introduced for mathematical reasoning, was extended to general reasoning \citep{deepseek2025r1}, showing that RL with rule-based outcome rewards alone sufficed to elicit complex reasoning behaviors including backtracking, self-questioning, and exploration of alternative solution paths. Theoretical analysis \citep{feng2024towards} showed that chain-of-thought lets bounded-depth transformers solve problems they otherwise could not. A growing body of work on inference-time compute control, including budget forcing \citep{muennighoff2025s1}, learned length control \citep{aggarwal2025l1}, and pause tokens \citep{goyal2024think}, further establishes extended thinking traces as the observable product of model deliberation.

\medskip

The separability of the thinking trace from the final answer opens a distinctive analytical opportunity. On the LSAT, where each question demands a structured chain of inference, the thinking trace is what reveals whether a model has reasoned its way to the answer, or merely arrived at it.

\section{Models}

Eight reasoning models are evaluated, spanning frontier and small scales, closed and open-source providers. All knowledge cutoffs precede April 2025, the release date of the Official Test, ensuring that no model could have encountered the primary evaluation data during training.

The two small models are reasoning-distilled from DeepSeek-R1 at roughly 1/100th the parameter count (Table~\ref{tab:models}). Because they are trained to imitate DeepSeek-R1's thinking traces, they inherit its \texttt{<think>...</think>} format and chain-of-thought style, providing a comparison along the scale dimension. They also serve as the base for the process supervision pilot in Section~\ref{sec:meth_prm}.

\begin{table}[h]
\caption{Models evaluated. Frontier models (top) and small models (bottom). Arch distinguishes dense transformers (all parameters active per token) from sparse Mixture-of-Experts (MoE) architectures, in which a router activates only a subset of experts per token. Total reports total parameter count; Active reports parameters activated per token. These are equal for dense models and differ for MoE models: Kimi K2 Thinking, for example, has 1T total parameters but activates only 32B per token. Parameter counts for the three closed models are not publicly disclosed.}
\label{tab:models}
\centering
\resizebox{\textwidth}{!}{%
\begin{tabular}{llcccccccc}
\toprule
Model & Provider & Arch & Total & Active & Type & Cutoff & Think Visible & Think Toggle \\
\midrule
GPT-5 & OpenAI & --- & --- & --- & Closed & Sep 2024 & Summarized & Yes \\
Claude Opus 4 & Anthropic & --- & --- & --- & Closed & Mar 2025 & Summarized & Yes \\
Gemini 2.5 Pro & Google & --- & --- & --- & Closed & Jan 2025 & Summarized & No \\
DeepSeek-R1 & DeepSeek & MoE & 671B & 37B & Open & Jul 2024 & Full trace & No \\
Kimi K2 Thinking & Moonshot AI & MoE & 1T & 32B & Open & Dec 2024 & Full trace & No \\
QwQ-32B & Alibaba/Qwen & Dense & 32B & 32B & Open & Nov 2024 & Full trace & No \\
\midrule
DeepSeek-R1 Distill 7B & DeepSeek & Dense & 7B & 7B & Open & Jul 2024 & Full trace & No \\
DeepSeek-R1 Distill Llama 8B & DeepSeek & Dense & 8B & 8B & Open & Dec 2023 & Full trace & No \\
\bottomrule
\end{tabular}%
}
\end{table}

\clearpage

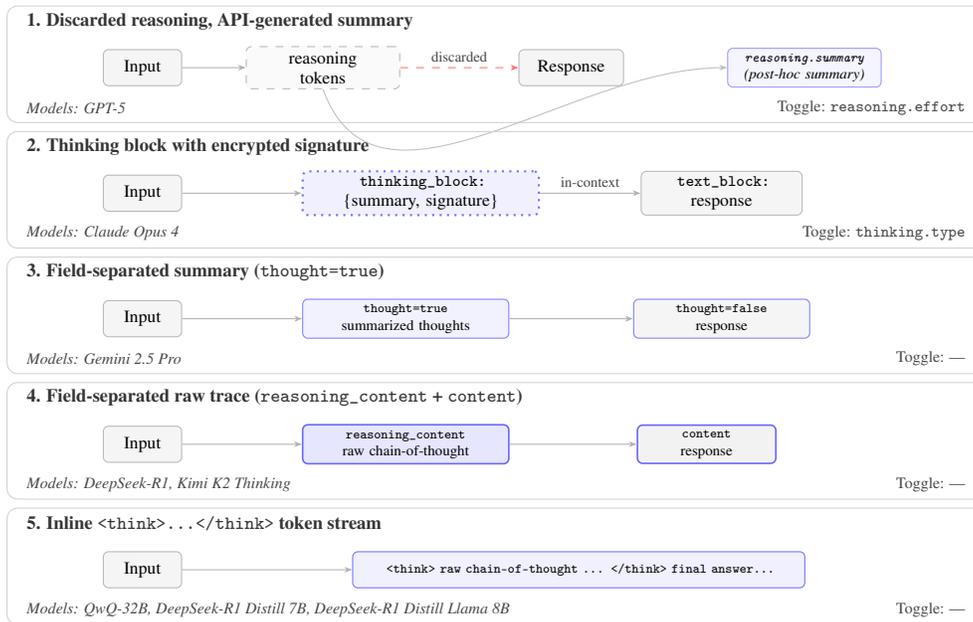
\begin{figure}[H]
\centering
\small
\begin{tikzpicture}[
  panel/.style={draw=gray!35, rounded corners=4pt, line width=0.4pt},
  phase/.style={draw, rounded corners=2pt, minimum height=0.48cm, align=center,
                font=\fontsize{6.5}{7.5}\selectfont, inner sep=2pt},
  inputbox/.style={phase, fill=gray!8, draw=gray!55, text width=0.9cm},
  respbox/.style={phase, fill=gray!8, draw=gray!55, text width=1.25cm},
  reasonHidden/.style={phase, fill=gray!4, draw=gray!55, dashed, text width=1.9cm},
  summaryBox/.style={phase, fill=blue!4, draw=blue!40, text width=1.9cm,
                     font=\fontsize{5.5}{6.5}\selectfont\itshape},
  thinkingBlock/.style={phase, fill=blue!6, draw=blue!55, dotted, thick, text width=3.0cm},
  thoughtsBox/.style={phase, fill=blue!5, draw=blue!50, text width=2.6cm,
                      font=\fontsize{5.5}{6.5}\selectfont},
  contentGemini/.style={phase, fill=gray!8, draw=blue!50, text width=2.2cm,
                        font=\fontsize{5.5}{6.5}\selectfont},
  reasonField/.style={phase, fill=blue!10, draw=blue!60, line width=0.6pt, text width=2.6cm,
                      font=\fontsize{5.5}{6.5}\selectfont},
  contentField/.style={phase, fill=gray!10, draw=blue!60, line width=0.6pt, text width=1.7cm,
                       font=\fontsize{5.5}{6.5}\selectfont},
  streamBox/.style={phase, fill=blue!6, draw=blue!55, text width=5.8cm,
                    font=\ttfamily\fontsize{5.5}{6.8}\selectfont, inner sep=3pt},
  arr/.style={-{Stealth[length=3pt]}, thin, gray!60},
  arrcut/.style={-{Stealth[length=3pt]}, dashed, thin, red!55},
  arrside/.style={-{Stealth[length=3pt]}, thin, gray!50},
  paneltitle/.style={font=\scriptsize\bfseries, anchor=west, text=black!80},
  modelnote/.style={font=\fontsize{6}{7}\selectfont\itshape, anchor=west, text=gray!55!black},
  togglenote/.style={font=\fontsize{6}{7}\selectfont, anchor=east, text=gray!55!black},
  annot/.style={font=\fontsize{5.5}{6.5}\selectfont, text=gray!55!black, align=center},
]

\def\pw{13}
\def\ph{1.55}
\def\pg{0.12}

\begin{scope}[yshift=0cm]
\draw[panel] (0, 0) rectangle (\pw, -\ph);
\node[paneltitle] at (0.15, -0.2) {\textbf{1.} Discarded reasoning, API-generated summary};
\node[inputbox] (p1in) at (1.8, -0.82) {Input};
\node[reasonHidden] (p1r) at (4.2, -0.82) {reasoning\\tokens};
\node[respbox] (p1out) at (7.5, -0.82) {Response};
\node[summaryBox] (p1sum) at (10.6, -0.82) {\texttt{reasoning.summary}\\(post-hoc summary)};
\draw[arr] (p1in) -- (p1r);
\draw[arrcut] (p1r) -- node[annot, above=-1pt, pos=0.5] {discarded} (p1out);
\draw[arrside] (p1r.south) to[out=-70, in=180] (p1sum.west);
\node[modelnote] at (0.15, -\ph+0.2) {Models: GPT-5};
\node[togglenote] at (\pw-0.15, -\ph+0.2) {Toggle: \texttt{reasoning.effort}};
\end{scope}

\begin{scope}[yshift=-1.67cm]
\draw[panel] (0, 0) rectangle (\pw, -\ph);
\node[paneltitle] at (0.15, -0.2) {\textbf{2.} Thinking block with encrypted signature};
\node[inputbox] (p2in) at (1.8, -0.82) {Input};
\node[thinkingBlock] (p2t) at (5.5, -0.82) {\texttt{thinking\_block:}\\\{summary, signature\}};
\node[respbox, text width=2.0cm] (p2out) at (9.5, -0.82) {\texttt{text\_block:}\\response};
\draw[arr] (p2in) -- (p2t);
\draw[arr] (p2t) -- node[annot, above=-1pt] {in-context} (p2out);
\node[modelnote] at (0.15, -\ph+0.2) {Models: Claude Opus 4};
\node[togglenote] at (\pw-0.15, -\ph+0.2) {Toggle: \texttt{thinking.type}};
\end{scope}

\begin{scope}[yshift=-3.34cm]
\draw[panel] (0, 0) rectangle (\pw, -\ph);
\node[paneltitle] at (0.15, -0.2) {\textbf{3.} Field-separated summary (\texttt{thought=true})};
\node[inputbox] (p3in) at (1.8, -0.82) {Input};
\node[thoughtsBox] (p3t) at (5.3, -0.82) {\texttt{thought=true}\\summarized thoughts};
\node[contentGemini] (p3out) at (9.5, -0.82) {\texttt{thought=false}\\response};
\draw[arr] (p3in) -- (p3t);
\draw[arr] (p3t) -- (p3out);
\node[modelnote] at (0.15, -\ph+0.2) {Models: Gemini 2.5 Pro};
\node[togglenote] at (\pw-0.15, -\ph+0.2) {Toggle: ---};
\end{scope}

\begin{scope}[yshift=-5.01cm]
\draw[panel] (0, 0) rectangle (\pw, -\ph);
\node[paneltitle] at (0.15, -0.2) {\textbf{4.} Field-separated raw trace (\texttt{reasoning\_content} + \texttt{content})};
\node[inputbox] (p4in) at (1.8, -0.82) {Input};
\node[reasonField] (p4r) at (5.3, -0.82) {\texttt{reasoning\_content}\\raw chain-of-thought};
\node[contentField] (p4c) at (9.3, -0.82) {\texttt{content}\\response};
\draw[arr] (p4in) -- (p4r);
\draw[arr] (p4r) -- (p4c);
\node[modelnote] at (0.15, -\ph+0.2) {Models: DeepSeek-R1, Kimi K2 Thinking};
\node[togglenote] at (\pw-0.15, -\ph+0.2) {Toggle: ---};
\end{scope}

\begin{scope}[yshift=-6.68cm]
\draw[panel] (0, 0) rectangle (\pw, -\ph);
\node[paneltitle] at (0.15, -0.2) {\textbf{5.} Inline \texttt{<think>...</think>} token stream};
\node[inputbox] (p5in) at (1.8, -0.82) {Input};
\node[streamBox] (p5s) at (7.6, -0.82) {<think> raw chain-of-thought ... </think> final answer...};
\draw[arr] (p5in) -- (p5s);
\node[modelnote] at (0.15, -\ph+0.2) {Models: QwQ-32B, DeepSeek-R1 Distill 7B, DeepSeek-R1 Distill Llama 8B};
\node[togglenote] at (\pw-0.15, -\ph+0.2) {Toggle: ---};
\end{scope}

\end{tikzpicture}

\caption{How the eight reasoning models expose their thinking traces. The models cluster into five architecturally distinct mechanisms that differ in generation, visibility, and multi-turn handling; see text for details.}
\label{fig:visibility}
\end{figure}

\begin{figure}[H]
\centering
\small

\newtcolorbox{inputbox}{
  enhanced,
  colback=white, colframe=gray!55, boxrule=0.5pt, arc=2pt,
  left=6pt, right=6pt, top=5pt, bottom=5pt,
  fontupper=\scriptsize\ttfamily,
  halign=flush left,
  title={\scriptsize\bfseries LSAT Official Test $\cdot$ April 2025 $\cdot$ Logical Reasoning $\cdot$ Q5},
  colbacktitle=gray!18, coltitle=black,
  fonttitle=\scriptsize\bfseries,
  attach boxed title to top left={yshift=-2pt, xshift=6pt},
  boxed title style={colback=gray!18, colframe=gray!55, boxrule=0.4pt, arc=1pt,
                     left=4pt, right=4pt, top=1.5pt, bottom=1.5pt}
}
\newtcolorbox{thinkbox}[1][]{
  colback=green!3, colframe=green!50!black, boxrule=0.6pt, arc=2pt,
  left=4pt, right=4pt, top=3pt, bottom=3pt, fontupper=\scriptsize\ttfamily, #1
}
\newtcolorbox{thinkboxweak}[1][]{
  colback=red!3, colframe=red!50!black, boxrule=0.6pt, arc=2pt,
  left=4pt, right=4pt, top=3pt, bottom=3pt, fontupper=\scriptsize\ttfamily, #1
}

\begin{inputbox}
Some scholars maintain that the first government other than China to issue any form of paper currency was the Massachusetts colony in 1690. But these scholars are incorrect. During a coin shortage in New France in 1685, playing cards were assigned an official value and issued as pay. These cards could be used to buy goods and were convertible to coins.\\[3pt]
\textbf{Which principle, if assumed, makes the conclusion properly drawn?}\\[2pt]
(A)~The playing cards were made of paper.\\
(B)~China did not issue paper currency in the 1600s.\\
(C)~Playing cards were not available in Massachusetts before 1690.\\
(D)~Massachusetts's issue was not prompted by a coin shortage.\\
(E)~Every government that issued paper currency left records.
\end{inputbox}

{\small\color{gray!70!black}Stimulus and answer choices abbreviated.}

\vspace{6pt}

\begin{thinkbox}[title={\scriptsize\bfseries Thinking Phase}]
The conclusion says ``paper currency.'' The evidence is about ``playing cards.'' For the conclusion to hold, we must assume that these playing cards count as paper currency. If the playing cards were \textbf{not made of paper}, they would not be paper currency, and the argument fails.\quad$\longrightarrow$\quad\textbf{Answer: (A) \color{green!50!black}\checkmark}
\end{thinkbox}

\begin{tikzpicture}
\node[anchor=west, font=\small, text=green!50!black] at (0, 0) {DeepSeek-R1 (671B), frontier.};
\end{tikzpicture}

\begin{thinkboxweak}[title={\scriptsize\bfseries Thinking Phase}]
Option C is saying that playing cards weren't available in Massachusetts before 1690. If that were true, then the researcher's point holds\ldots{} So yes, this seems like a necessary assumption. \textit{[Misses the core gap---``playing cards'' $\neq$ ``paper currency'' unless made of paper---and selects an irrelevant assumption.]}\quad$\longrightarrow$\quad\textbf{Answer: (C) \color{red!50!black}$\boldsymbol{\times}$}
\end{thinkboxweak}

\begin{tikzpicture}
\node[anchor=west, font=\small, text=red!50!black] at (0, 0) {DS-R1 Distill 7B, small (distilled from DeepSeek-R1).};
\end{tikzpicture}

\caption{Reasoning form vs.\ reasoning quality on a Logical Reasoning question from the Official Test (correct answer: A). Stimulus, answer choices, and thinking traces abbreviated for space; see Appendix~\ref{app:traces} for the full question and complete traces. DeepSeek-R1 (top) identifies the necessary assumption. Its distilled 7B counterpart (bottom) selects the wrong answer, missing the critical logical bridge.}
\label{fig:formquality}
\end{figure}

\clearpage

All eight models generate an internal chain of thought before producing a final answer. This sequence of intermediate steps is hereafter referred to as the \emph{thinking trace}. The mechanism by which the trace is exposed to the user, however, differs substantially across providers. Figure~\ref{fig:visibility} groups the eight models into five architecturally distinct mechanisms. \textbf{1.}~GPT-5 discards its reasoning tokens from context after the reasoning phase; only a post-hoc summary (\texttt{reasoning.summary}) is returned to the user, and a \texttt{reasoning\_effort} parameter controls the depth \citep{openai2025reasoningapi}. \textbf{2.}~Claude Opus 4 emits thinking as discrete content blocks alongside text blocks, with an encrypted \texttt{signature} field that allows the full thinking to be carried across turns \citep{anthropic2025extendedthinking}. \textbf{3.}~Gemini 2.5 Pro returns summarized thoughts in response parts marked \texttt{thought=true}, structurally separate from the \texttt{thought=false} parts that carry the response \citep{google2025geminithinking}. \textbf{4.}~DeepSeek-R1 \citep{deepseek2025api} and Kimi K2 Thinking \citep{moonshot2025api} return the raw chain-of-thought in a dedicated \texttt{reasoning\_content} API field, structurally distinct from the \texttt{content} field that carries the response. \textbf{5.}~QwQ-32B \citep{qwen2025qwqhf} and the two DeepSeek-R1 Distill variants \citep{deepseek2025distillhf, deepseek2025distillhfllama} emit raw \texttt{<think>...</think>} tags directly inside the output token stream, with no API-level separation between thinking and response. These five mechanisms also differ in what persists across turns: GPT-5 drops its reasoning; Claude Opus 4 carries the full thinking forward via encrypted signatures, while Gemini 2.5 Pro retains only the returned summary in conversation text; DeepSeek-R1 requires stripping the reasoning field before the next request while Kimi K2 Thinking requires keeping it; QwQ-32B and the distilled models drop the thinking when constructing chat history.

Figure~\ref{fig:formquality} illustrates this quality gap on an actual LSAT question, contrasting a frontier model that identifies the critical logical gap with a small model that misses it entirely. Complete thinking traces and responses for all eight models on this question are reproduced in Appendix~\ref{app:traces}.
\section{Datasets}
\label{sec:datasets}

All analyses group questions into two categories, LR (Logical Reasoning) and RC (Reading Comprehension).

\textbf{Test set.} Each year, the Law School Admission Council (LSAC) publicly discloses one previously administered LSAT \citep{lsac2025disclosed}. The most recent disclosure, the April 2025 administration, postdates the knowledge cutoff of every model in this study, providing a test set free of contamination risk. It comprises 77 questions (50 LR, 27 RC). Hereafter referred to as the \emph{Official Test}.

\textbf{Validation set.} Ten consecutive tests (PrepTests 150--159) comprising 1,037 questions (633 LR, 404 RC). This larger dataset is used for statistical power.

\textbf{Training set.} Two tests (PrepTests 140--141) comprising 209 questions (128 LR, 81 RC) with official LSAC per-choice explanations, further reviewed by a human expert who scored a perfect 180 on the LSAT. This dataset is used exclusively for process reward model training in Section~\ref{sec:prm}. No overlap exists with the Official Test or the validation set.

Throughout all experiments, questions are presented individually, one per API call, with no cross-question context even for Reading Comprehension questions sharing the same passage.

\section{Methodology}

Prior work has shown that language models can be sensitive to prompt phrasing, answer-position artifacts, and variation across repeated samples. Reasoning models, with their dedicated thinking phase, may behave differently. Sections~\ref{sec:meth_prompt}--\ref{sec:meth_sc} test whether these factors influence performance at either the frontier or small scale. Section~\ref{sec:meth_ablation} isolates the role of thinking. Section~\ref{sec:meth_prm} explores whether thinking quality can be externally improved via a process reward model.

\subsection{Prompt Sensitivity}
\label{sec:meth_prompt}
Language model performance can vary substantially with prompt phrasing, a phenomenon documented across reasoning benchmarks, where even minor wording changes shift accuracy by several percentage points \citep{zhao2021calibrate, lu2022fantastically}. To test whether reasoning models exhibit similar sensitivity, all eight models are evaluated under three conditions.

\textbf{Condition A (Minimal Prompting).} No system prompt. User message: \texttt{``Here is an LSAT [Logical Reasoning / Reading Comprehension] question. What is the answer?''} followed by the question text and five answer choices.

\textbf{Condition B (Structured Prompting).} System prompt assigns an expert persona (an LSAT instructor with a perfect 180) and prescribes a structured chain-of-thought methodology, identifying premises and conclusions, eliminating each wrong answer with explicit reasoning, and confirming the final selection.

\textbf{Condition C (Constrained Prompting).} System prompt: \texttt{``Respond with ONLY the letter of the correct answer (A, B, C, D, or E). Nothing else.''}

For Conditions A and B, answers are extracted via a lightweight parser that identifies the selected letter without influencing model output. Cochran's Q test compares accuracy across the three conditions; TOST equivalence testing ($\pm$2pp margin) is applied to pairwise comparisons on PT 150--159. Full prompt templates are provided in Appendix~\ref{app:prompts}.

\subsection{Position Bias}
\label{sec:meth_position}

Language models have been shown to exhibit position bias in multiple-choice settings, preferring answers placed in certain positions, typically early options such as (A) or late options such as (E), independent of content \citep{pezeshkpour2023large, zheng2024large}. To test whether reasoning models are susceptible to this artifact, shuffled variants of both datasets (fixed seed across models) permute answer choices. McNemar's exact test compares paired accuracy between original and shuffled orderings; $\chi^2$ goodness-of-fit tests letter-selection distributions against uniform. TOST equivalence testing ($\pm$2pp) is applied at the section level.

\subsection{Self-Consistency}
\label{sec:meth_sc}

Self-consistency \citep{wang2023selfconsistency} improves language model accuracy by sampling multiple reasoning paths and selecting the most common answer via majority vote, exploiting the assumption that correct reasoning is more likely to converge than incorrect reasoning. To test whether this strategy yields meaningful gains for reasoning models, or whether their errors are too deterministic for majority voting to correct, each model generates five independent responses per question, following the smallest sample size evaluated in \citep{wang2023selfconsistency}. Self-consistency accuracy (SC@5) is computed via majority vote and compared against pass@1 (the first generated response) using McNemar's exact test. Unanimity, the proportion of questions receiving 5/5 identical answers, is reported as a measure of response determinism.

\subsection{Ablation Study: Thinking}
\label{sec:meth_ablation}

The preceding experiments vary the input and decoding strategy. This experiment targets the reasoning process itself. To isolate the contribution of the thinking phase, GPT-5 (\texttt{reasoning\_effort: none}) and Claude Opus 4 (\texttt{thinking: disabled}) are evaluated with thinking turned off, treating the API toggle as a binary ablation of the reasoning component. These are the only two models whose APIs permit disabling thinking. Gemini 2.5 Pro does not expose a toggle, and the open-source models do not provide a non-thinking inference mode. McNemar's exact test compares thinking-on vs.\ thinking-off accuracy. Cohen's $h$ quantifies effect sizes. Results are broken down by section (LR vs.\ RC) to test whether thinking differentially benefits logical inference.

\subsection{Pilot Study: Process Reward Modeling}
\label{sec:meth_prm}

The four preceding experiments target prompting, answer ordering, sample diversity, and the presence of a thinking phase, all of which apply uniformly across frontier and small scales. What none of them probes directly is the rigor of the reasoning carried out within the trace itself, the property that distinguishes a model that has reasoned its way to the answer from one that has merely arrived at it. This pilot study tests whether such rigor can be externally identified and selected for, by training a process reward model \citep{uesato2022solving, lightman2023lets} on official LSAT explanations and using its scores to perform Best-of-$N$ selection at inference time. The base model is DeepSeek-R1 Distill 7B, the smallest model in the evaluation and the scale most amenable to fine-tuning on a single GPU.

\textbf{Training signal construction.} The training signal must score the thinking trace itself rather than the final answer letter, since whether the trace reaches the right letter and whether the trace reasons its way there are independent properties. The official LSAT explanations for PT 140 and PT 141, reviewed by a human expert, provide the supervision, since they label each of the five answer choices as correct or incorrect and give the underlying logical reason for that label. Each thinking trace is then graded against these explanations choice by choice through a two-pass procedure. In the first pass, the judgment that the trace expresses about each answer choice is classified as correct, incorrect, or not addressed, and one point is awarded when this judgment agrees with the official label. Choices the trace fails to address receive zero points by construction. In the second pass, for every choice on which the trace agreed with the official judgment, the reasoning that the trace offers for that choice is compared against the official explanation, and one further point is awarded when the trace captures the same underlying logical reason rather than aligning with the official label only nominally. The per-trace score sums across the five answer choices on a 0 to 10 scale and is then normalized to the unit interval $[0, 1]$, so that a score of 1.0 requires the correct judgment together with the matching reason on every one of the five answer choices. Empirically, this score averaged 0.89 on traces leading to a correct final answer and 0.47 on those leading to an incorrect one.

\textbf{Self-PRM training.} The same base model is fine-tuned as its own PRM via QLoRA (4-bit quantization, LoRA $r = 8$), using all 209 training examples from PT 140 and PT 141. Given a question and a thinking trace, the model learns to predict the reasoning rigor score. This self-PRM design avoids the cross-model distribution mismatch that arises when a verifier trained on the reasoning patterns of one model is applied to another \citep{park2025calibration}. The full training configuration is provided in Appendix~\ref{app:prm}.

\textbf{Inference.} At test time, the generator produces five candidate responses per question. The PRM scores each thinking trace and selects the highest-scoring response (Best-of-5). Under the same compute budget of five samples per question, this is compared against pass@1 (the first generated response) and SC@5 (majority vote) on the held-out Official Test.

\subsection{Statistical Framework}

All statistical tests are two-tailed with $\alpha = 0.05$. McNemar's exact test is used for paired binary comparisons, Cochran's Q for three-condition comparisons, TOST ($\pm$2pp margin) for equivalence, and Cohen's $h$ for effect sizes. TOST is applied on PT 150--159, where the sample size ($N = 1{,}037$) provides sufficient power for the $\pm$2pp equivalence margin. On the Official Test ($N = 77$), the minimum detectable equivalence margin exceeds $\pm$2pp, so only significance tests are reported.

\section{Results}
\label{sec:results}

Table~\ref{tab:main} establishes the baseline for all subsequent experiments, reporting the accuracy of each model under Condition A (Minimal Prompting) with no answer reordering or sampling strategies. Kimi K2 Thinking achieved a perfect score under this condition, and DeepSeek-R1 matched it under Conditions B and C (Table~\ref{tab:prompt_160}). These are the first documented instances of a language model achieving a perfect score on an officially disclosed LSAT. The remaining frontier models scored 92--99\% across both datasets. The two small models, included as the base for the process supervision pilot in Section~\ref{sec:prm}, scored below 64\%.

\begin{table}[h]
\caption{Baseline accuracy under default conditions (no system prompt, no answer reordering, single sample). LR = Logical Reasoning, RC = Reading Comprehension. Frontier models (top) and small models (bottom).}
\label{tab:main}
\centering
\small
\begin{tabular}{lcccccc}
\toprule
& \multicolumn{3}{c}{Official Test (N=77)} & \multicolumn{3}{c}{PT 150--159 (N=1037)} \\
\cmidrule(lr){2-4} \cmidrule(lr){5-7}
Model & LR & RC & Total & LR & RC & Total \\
\midrule
GPT-5 & 98.0 & 100.0 & 98.7 & 99.1 & 98.5 & 98.8 \\
Claude Opus 4 & 98.0 & 100.0 & 98.7 & 98.0 & 97.8 & 97.9 \\
Gemini 2.5 Pro & 94.0 & 100.0 & 96.1 & 99.1 & 97.8 & 98.6 \\
DeepSeek-R1 & 100.0 & 96.3 & 98.7 & 97.8 & 97.5 & 97.7 \\
Kimi K2 Thinking & 100.0 & 100.0 & 100.0 & 97.8 & 96.3 & 97.2 \\
QwQ-32B & 92.0 & 92.6 & 92.2 & 92.9 & 93.8 & 93.2 \\
\midrule
DeepSeek-R1 Distill 7B & 54.0 & 59.3 & 55.8 & 54.0 & 60.9 & 56.7 \\
DeepSeek-R1 Distill Llama 8B & 46.0 & 66.7 & 53.2 & 57.0 & 72.8 & 63.2 \\
\bottomrule
\end{tabular}
\end{table}

\subsection{Prompt Sensitivity}

\begin{table}[h]
\caption{Prompt sensitivity on the Official Test ($N = 77$). Cochran's Q across three conditions. Per-section breakdowns in Table~\ref{tab:prompt_section_160}.}
\label{tab:prompt_160}
\centering
\small
\begin{tabular}{lccccc}
\toprule
Model & Cond A & Cond B & Cond C & $Q(2)$ & $p$ \\
\midrule
GPT-5 & 98.7 & 98.7 & 98.7 & 0.00 & 1.000 \\
Claude Opus 4 & 98.7 & 96.1 & 96.1 & 2.67 & .264 \\
Gemini 2.5 Pro & 96.1 & 96.1 & 98.7 & 2.67 & .264 \\
DeepSeek-R1 & 98.7 & 100.0 & 100.0 & 2.00 & .368 \\
Kimi K2 Thinking & 100.0 & 98.7 & 96.1 & 3.50 & .174 \\
QwQ-32B & 92.2 & 93.5 & 93.5 & 0.40 & .819 \\
\midrule
DS-R1 Distill 7B & 55.8 & 50.6 & 50.6 & 1.00 & .607 \\
DS-R1 Distill Llama 8B & 53.2 & 59.7 & 70.1 & 9.21 & .010 \\
\bottomrule
\end{tabular}
\end{table}

On the Official Test, no model shows a significant difference across prompt conditions (all Cochran's Q $p > .05$). DS-R1 Distill Llama 8B is the sole exception ($Q = 9.21$, $p = .010$), but as the larger dataset confirms below, this does not replicate.

\begin{table}[h]
\caption{Prompt sensitivity on PT 150--159 ($N = 1{,}037$). Cochran's Q test across three conditions (A/B/C). Per-section breakdowns in Table~\ref{tab:prompt_section}.}
\label{tab:prompt}
\centering
\small
\begin{tabular}{lcccccl}
\toprule
Model & Cond A & Cond B & Cond C & $Q(2)$ & $p$ & TOST \\
\midrule
GPT-5 & 98.8 & 98.6 & 98.6 & 0.93 & .627 & 9/9 EQUIV \\
Claude Opus 4 & 97.9 & 98.0 & 97.8 & 0.22 & .895 & 9/9 EQUIV \\
Gemini 2.5 Pro & 98.6 & 97.8 & 98.2 & 3.56 & .169 & 8/9 EQUIV \\
DeepSeek-R1 & 97.7 & 97.7 & 97.9 & 0.36 & .834 & 8/9 EQUIV \\
Kimi K2 Thinking & 97.2 & 96.5 & 97.0 & 1.28 & .528 & 5/9 EQUIV \\
QwQ-32B & 93.2 & 93.3 & 93.2 & 0.07 & .967 & 4/9 EQUIV \\
\midrule
DS-R1 Distill 7B & 56.7 & 57.9 & 56.6 & 0.79 & .674 & 0/9 \\
DS-R1 Distill Llama 8B & 63.2 & 61.7 & 62.7 & 0.96 & .617 & 0/9 \\
\bottomrule
\end{tabular}
\end{table}

The larger PT 150--159 dataset ($N = 1{,}037$) confirms this result across all eight models (all Cochran's Q $p > .05$), and the apparent prompt sensitivity for DS-R1 Distill Llama 8B does not replicate ($p = .617$). TOST equivalence testing establishes that the differences are not merely non-significant but negligibly small. GPT-5 and Claude Opus 4 achieved 9/9 pairwise equivalence within $\pm$2pp, and Gemini 2.5 Pro and DeepSeek-R1 achieved 8/9. Small models were likewise non-significant on Cochran's Q but could not establish equivalence (0/9 TOST), reflecting higher variability at lower accuracy levels. The one section-level exception is Gemini 2.5 Pro LR ($Q = 9.50$, $p = .009$, $h = 0.124$), where Condition B (Structured Prompting) yielded lower accuracy than the other two conditions. Section-level breakdowns (Tables~\ref{tab:prompt_section_160}--\ref{tab:prompt_section}) confirm that the pattern otherwise holds uniformly across LR and RC.

Frontier models maintain 97--99\% accuracy regardless of whether they receive no prompt, an expert persona, or a constrained format. Prompt design is not a lever for improving, or degrading, reasoning model performance on the LSAT.

\subsection{Position Bias}

\begin{table}[h]
\caption{Position bias on the Official Test ($N = 77$). Original vs.\ shuffled (McNemar's exact test). $b$ = original-only correct; $c$ = shuffled-only correct; $h$ = Cohen's $h$. Per-section breakdowns in Table~\ref{tab:shuffle_section_160}.}
\label{tab:shuffle_160}
\centering
\small
\begin{tabular}{lcccccc}
\toprule
Model & Original & Shuffled & $b$ & $c$ & $p$ & $h$ \\
\midrule
GPT-5 & 98.7 & 97.4 & 1 & 0 & 1.000 & $-$0.095 \\
Claude Opus 4 & 98.7 & 98.7 & 1 & 1 & 1.000 & 0.000 \\
Gemini 2.5 Pro & 96.1 & 97.4 & 0 & 1 & 1.000 & +0.074 \\
DeepSeek-R1 & 98.7 & 98.7 & 1 & 1 & 1.000 & 0.000 \\
Kimi K2 Thinking & 100.0 & 98.7 & 1 & 0 & 1.000 & $-$0.228 \\
QwQ-32B & 92.2 & 94.8 & 1 & 3 & .625 & +0.106 \\
\midrule
DS-R1 Distill 7B & 55.8 & 49.4 & 17 & 12 & .458 & $-$0.130 \\
DS-R1 Distill Llama 8B & 53.2 & 64.9 & 6 & 15 & .078 & +0.238 \\
\bottomrule
\end{tabular}
\end{table}

Shuffling answer-choice order produces no significant accuracy change for any model on the Official Test (all McNemar $p > .05$), though the small sample ($N = 77$) limits statistical power.

\begin{table}[h]
\caption{Position bias on PT 150--159 ($N = 1{,}037$). Original vs.\ shuffled (McNemar's exact test). Per-section breakdowns in Table~\ref{tab:shuffle_section}.}
\label{tab:shuffle}
\centering
\small
\begin{tabular}{lccccccl}
\toprule
Model & Original & Shuffled & $b$ & $c$ & $p$ & $h$ & TOST \\
\midrule
GPT-5 & 98.8 & 98.7 & 3 & 2 & 1.000 & +0.009 & 3/3 EQUIV \\
Claude Opus 4 & 97.9 & 98.0 & 5 & 6 & 1.000 & $-$0.007 & 3/3 EQUIV \\
Gemini 2.5 Pro & 98.6 & 98.5 & 4 & 3 & 1.000 & +0.008 & 3/3 EQUIV \\
DeepSeek-R1 & 97.7 & 97.4 & 14 & 11 & .689 & +0.019 & 2/3 EQUIV \\
Kimi K2 Thinking & 97.2 & 97.0 & 17 & 15 & .860 & +0.012 & 1/3 EQUIV \\
QwQ-32B & 93.2 & 93.2 & 32 & 32 & 1.000 & 0.000 & 2/3 EQUIV \\
\midrule
DS-R1 Distill 7B & 56.7 & 57.5 & 150 & 158 & .690 & $-$0.016 & 0/3 \\
DS-R1 Distill Llama 8B & 63.2 & 63.1 & 130 & 129 & 1.000 & +0.002 & 0/3 \\
\bottomrule
\end{tabular}
\end{table}

PT 150--159 confirms the absence of position bias across all eight models (all $|h| < 0.1$). TOST equivalence testing establishes that the differences are negligibly small, with GPT-5, Claude Opus 4, and Gemini 2.5 Pro achieving equivalence within $\pm$2pp on all three section-level tests (3/3 EQUIV). $\chi^2$ goodness-of-fit tests on answer-letter distributions (A--E) versus uniform were non-significant across all comparisons. Small models likewise showed no systematic bias ($p > .5$, $|h| < 0.02$), though TOST could not establish equivalence, reflecting higher variability at lower accuracy levels.

The discordant pair counts reveal an important asymmetry. Frontier models changed only 0.5--6.2\% of items between original and shuffled conditions, while small models changed 25--30\%, but symmetrically ($b \approx c$), indicating stochastic rather than position-driven responding. Frontier accuracy is content-driven and stable across answer orderings. Section-level breakdowns (Tables~\ref{tab:shuffle_section_160}--\ref{tab:shuffle_section}) confirm that this holds for both LR and RC independently.

\subsection{Self-Consistency}

\begin{table}[h]
\caption{Self-consistency on the Official Test ($N = 77$). pass@1 = first of five samples; SC@5 = majority vote. $b$ = pass@1-only correct; $c$ = SC@5-only correct. Unanimity = proportion of questions with 5/5 identical answers. Per-section breakdowns in Table~\ref{tab:sc_section_160}.}
\label{tab:sc_160}
\centering
\small
\begin{tabular}{lccccccc}
\toprule
Model & pass@1 & SC@5 & $\Delta$ & $b$ & $c$ & $p$ & Unanimity \\
\midrule
GPT-5 & 97.4 & 97.4 & 0.0 & 0 & 0 & 1.000 & 97.4\% \\
Claude Opus 4 & 93.5 & 96.1 & +2.6 & 0 & 2 & .500 & 94.8\% \\
Gemini 2.5 Pro & 97.4 & 97.4 & 0.0 & 0 & 0 & 1.000 & 97.4\% \\
DeepSeek-R1 & 98.7 & 98.7 & 0.0 & 0 & 0 & 1.000 & 97.4\% \\
Kimi K2 Thinking & 97.4 & 98.7 & +1.3 & 0 & 1 & 1.000 & 94.8\% \\
QwQ-32B & 93.5 & 96.1 & +2.6 & 0 & 2 & .500 & 92.2\% \\
\midrule
DS-R1 Distill 7B & 51.9 & 55.8 & +3.9 & 7 & 10 & .629 & 22.1\% \\
DS-R1 Distill Llama 8B & 55.8 & 59.7 & +3.9 & 3 & 6 & .508 & 37.7\% \\
\bottomrule
\end{tabular}
\end{table}

Frontier unanimity rates on the Official Test are 92--97\%, and no comparison reaches significance. The few frontier errors are reproduced identically across all five samples, leaving no disagreement for majority voting to resolve. The larger dataset quantifies this pattern with greater precision.

\begin{table}[h]
\caption{Self-consistency on PT 150--159 ($N = 1{,}037$). McNemar's exact test. Per-section breakdowns in Table~\ref{tab:sc_section}.}
\label{tab:sc}
\centering
\small
\begin{tabular}{lcccccccc}
\toprule
Model & pass@1 & SC@5 & $\Delta$ & $b$ & $c$ & $p$ & $h$ & Unanimity \\
\midrule
GPT-5 & 97.3 & 99.0 & +1.7 & 0 & 18 & $<$.001 & +0.133 & 92.8\% \\
Claude Opus 4 & 97.1 & 98.0 & +0.9 & 4 & 13 & .049 & +0.056 & 91.7\% \\
Gemini 2.5 Pro & 97.6 & 98.6 & +1.1 & 3 & 14 & .013 & +0.079 & 93.7\% \\
DeepSeek-R1 & 97.1 & 97.8 & +0.7 & 2 & 9 & .065 & +0.043 & 94.9\% \\
Kimi K2 Thinking & 97.2 & 98.4 & +1.2 & 2 & 14 & .004 & +0.079 & 92.8\% \\
QwQ-32B & 94.5 & 94.9 & +0.4 & 7 & 11 & .481 & +0.017 & 88.9\% \\
\midrule
DS-R1 Distill 7B & 58.7 & 62.6 & +3.9 & 58 & 98 & .002 & +0.079 & 35.3\% \\
DS-R1 Distill Llama 8B & 63.9 & 68.9 & +4.9 & 48 & 99 & $<$.001 & +0.104 & 44.6\% \\
\bottomrule
\end{tabular}
\end{table}

No frontier model shows a practically meaningful improvement from self-consistency on PT 150--159. Several reach statistical significance (GPT-5 $p < .001$, Kimi K2 Thinking $p = .004$, Gemini 2.5 Pro $p = .013$), but all effect sizes are negligible ($h < 0.14$, all $\Delta < 2$pp). QwQ-32B ($p = .481$) and DeepSeek-R1 ($p = .065$) do not reach significance even at $N = 1{,}037$. The statistical significance at the overall level reflects the power of a large dataset to detect differences too small to matter in practice.

Frontier unanimity rates (89--95\%) explain why. The wrong answer on missed questions is typically chosen 5/5 times, leaving no disagreement for majority voting to exploit. GPT-5 illustrates the point with $b = 0$, where self-consistency never converted a correct first answer to an incorrect one, yet recovered only 18 of 1,037 questions. Small models show the opposite pattern, with unanimity at 35--45\% and SC@5 gains of 3.9--4.9pp, but these gains come with substantial churn (DS-R1 Distill 7B gained 98 questions but lost 58).

Per-section breakdowns are reported in Tables~\ref{tab:sc_section_160}--\ref{tab:sc_section}. Frontier errors on the LSAT are deterministic and resistant to sampling-based correction. Closing the remaining gap requires not more samples but more rigorous reasoning, a hypothesis the next two sections test directly.

\subsection{Ablation Study: Thinking}
 
\begin{table}[h]
\caption{Thinking ablation on the Official Test ($N = 77$). McNemar's exact test. $b$ = think-ON-only correct; $c$ = think-OFF-only correct.}
\label{tab:ablation_160}
\centering
\small
\begin{tabular}{llccccccc}
\toprule
Model & Section & ON & OFF & $\Delta$ & $b$ & $c$ & $p$ & $h$ \\
\midrule
\multirow{3}{*}{GPT-5} & Overall & 98.7 & 93.5 & +5.2 & 4 & 0 & .125 & +0.287 \\
& LR & 98.0 & 92.0 & +6.0 & 3 & 0 & .250 & +0.290 \\
& RC & 100.0 & 96.3 & +3.7 & 1 & 0 & 1.000 & +0.387 \\
\midrule
\multirow{3}{*}{Claude Opus 4} & Overall & 98.7 & 90.9 & +7.8 & 6 & 0 & .031 & +0.384 \\
& LR & 98.0 & 88.0 & +10.0 & 5 & 0 & .062 & +0.424 \\
& RC & 100.0 & 96.3 & +3.7 & 1 & 0 & 1.000 & +0.387 \\
\bottomrule
\end{tabular}
\end{table}
 
On the Official Test, disabling thinking reduced accuracy substantially for both models. Notably, $c = 0$ across all six comparisons, meaning thinking never converted a correct answer to an incorrect one. The larger dataset confirms the effect with greater statistical power.
 
\begin{table}[h]
\caption{Thinking ablation on PT 150--159 ($N = 1{,}037$). McNemar's exact test.}
\label{tab:ablation}
\centering
\small
\begin{tabular}{llccccccc}
\toprule
Model & Section & ON & OFF & $\Delta$ & $b$ & $c$ & $p$ & $h$ \\
\midrule
\multirow{3}{*}{GPT-5} & Overall & 98.8 & 95.5 & +3.4 & 38 & 3 & $<$.001 & +0.214 \\
& LR & 99.1 & 95.3 & +3.8 & 26 & 2 & $<$.001 & +0.244 \\
& RC & 98.5 & 95.8 & +2.7 & 12 & 1 & .003 & +0.169 \\
\midrule
\multirow{3}{*}{Claude Opus 4} & Overall & 97.9 & 95.4 & +2.5 & 33 & 7 & $<$.001 & +0.141 \\
& LR & 97.9 & 94.0 & +3.9 & 28 & 3 & $<$.001 & +0.207 \\
& RC & 97.8 & 97.5 & +0.2 & 5 & 4 & 1.000 & +0.016 \\
\bottomrule
\end{tabular}
\end{table}
 
Extended thinking significantly improved accuracy for both models (GPT-5: $+$3.4pp, $p < .001$; Claude Opus 4: $+$2.5pp, $p < .001$). The effect was strongly asymmetric: thinking-enabled models answered 38 (GPT-5) and 33 (Claude Opus 4) additional questions correctly while losing only 3 and 7 respectively.
 
The most notable finding is the section-level dissociation. The entire thinking benefit for Claude Opus 4 concentrated in Logical Reasoning ($+$3.9pp, $p < .001$) with effectively zero effect on Reading Comprehension ($+$0.2pp, $p = 1.0$). GPT-5 showed a similar but less extreme pattern (LR: $+$3.8pp; RC: $+$2.7pp). This pattern suggests that extended thinking benefits logical inference more than text comprehension.
 
While Cohen's $h$ values (0.14--0.24 on PT 150--159) indicate small effects by conventional standards, at near-ceiling accuracy ($>$95\%), a 3.4pp improvement eliminates approximately half of remaining errors.
 
\subsection{Pilot Study: Process Reward Modeling}
\label{sec:prm}

\begin{table}[h]
\caption{Process supervision vs.\ baselines on DeepSeek-R1 Distill 7B (Official Test, unseen during PRM training). LR = Logical Reasoning ($N = 50$), RC = Reading Comprehension ($N = 27$).}
\label{tab:prm_7b}
\centering
\small
\begin{tabular}{lccc}
\toprule
& \multicolumn{3}{c}{Accuracy (\%)} \\
\cmidrule(lr){2-4}
Method & LR & RC & Overall \\
\midrule
Baseline (pass@1) & 52.0 & 51.9 & 51.9 \\
SC@5 & 50.0 & 66.7 & 55.8 \\
PRM + BoN@5 & 62.0 & 66.7 & 63.6 \\
\midrule
$\Delta$ (PRM vs.\ pass@1) & +10.0 & +14.8 & +11.7 \\
$\Delta$ (PRM vs.\ SC) & +12.0 & 0.0 & +7.8 \\
\bottomrule
\end{tabular}
\end{table}

PRM-guided Best-of-5 selection outperforms both baselines, raising overall accuracy by 11.7 percentage points over pass@1 and 7.8 percentage points over self-consistency. The per-section breakdown is notable. Gains over self-consistency concentrate entirely in Logical Reasoning (+12.0pp) with zero improvement in Reading Comprehension. The same per-section pattern appears in the thinking ablation (Section~\ref{sec:meth_ablation}), where the thinking benefit for Claude Opus 4 concentrated in LR with near-zero effect on RC.

Two additional observations merit note. First, SC@5 slightly decreases LR accuracy relative to pass@1 (50.0\% vs.\ 52.0\%), which may reflect a correlated-error problem, where most samples make the same logical error and majority voting amplifies rather than corrects it. PRM-guided selection avoids this failure mode by evaluating reasoning rigor rather than answer frequency. Second, manual inspection of the nine questions where PRM succeeds but SC fails reveals that the PRM consistently identifies minority samples (1--2 out of 5) that engage more carefully with each answer choice, mirroring official LSAT explanation structure, while the majority converge on an incorrect answer through similar reasoning errors.

The pilot results are preliminary, conducted on a single model with 209 training examples, but they suggest that this kind of external supervision can improve small-model reasoning quality, motivating further work with larger training sets and additional model architectures.
\section{Discussion}

\subsection{A Perfect Score}

In early 2023, GPT-3.5 scored 149 on the LSAT \citep{openai2023gpt4}. Later that year, GPT-4 scored 163 \citep{openai2023gpt4}. Less than two years later, in this study, GPT-5 misses only one of 77 questions on the Official Test, a score that still maps to a perfect 180 on the LSAT scoring curve \citep{lsac2025disclosed}, and Kimi K2 Thinking goes one further by answering every question correctly for an unambiguous 180. Five of the six frontier models in this study exceed 97\% across over 1,000 questions.

This trajectory is not driven by scaling alone. Disabling the thinking phase reduces frontier accuracy by 3--8pp (Section~\ref{sec:meth_ablation}), establishing that the thinking phase itself contributes substantively to frontier performance. A perfect LSAT score is not simply the product of a larger model. It is the product of one that reasons.

\subsection{Reasoning Rigor as the Differentiator}

Three factors widely documented to shift language model performance prove inert on the LSAT. Prompt phrasing leaves frontier accuracy unchanged across minimal, structured, and constrained conditions (Section~\ref{sec:meth_prompt}). Shuffling answer choices reveals no position bias at the frontier (Section~\ref{sec:meth_position}). Self-consistency recovers only a handful of errors for frontier models, whose mistakes are reproduced identically across samples (Section~\ref{sec:meth_sc}). The standard toolkit for moving language model performance, in short, has little purchase on reasoning models at the top of the LSAT.

What does move frontier accuracy is the thinking phase itself. Disabling thinking in GPT-5 and Claude Opus 4 reduces accuracy by 3--8pp, with the drop concentrating in logical reasoning (Section~\ref{sec:meth_ablation}). Thinking is therefore doing substantive work on the LSAT, not merely adding latency or formatting structure. But it cannot be the only variable that matters, because the two distilled small models in Table~\ref{tab:main} already produce full thinking traces in the same \texttt{<think>...</think>} format and still score below 64\%. The presence of a thinking phase is therefore necessary but not sufficient, and the quality of the reasoning the trace carries matters at least as much as whether a trace is produced at all.

Section~\ref{sec:prm} independently shows that improving reasoning quality via PRM selection lifts small-model accuracy by +11.7pp, with gains again concentrated in logical reasoning. Because the two experiments operate at different scales and via different interventions, the convergence should be read as a consistent empirical pattern rather than a direct causal demonstration.

\subsection{Limitations and Future Directions}

This study evaluates models at a single point in time, and rapid model development may quickly alter the performance landscape. The thinking ablation was possible for only two models,  GPT-5 and Claude Opus 4, due to API constraints, limiting the generalizability of that finding. The primary test set (the Official Test, $N = 77$) provides limited statistical power for per-section analyses. The validation set (PT 150--159) provides greater power but predates all knowledge cutoffs and may therefore have been seen during pretraining. If a model had memorized these questions, it would score higher on PT 150--159 than on the post-cutoff Official Test, which no model could have encountered. For all frontier models and DS-R1 Distill 7B, the gap between the two datasets is within $\pm$2.5pp, consistent with sampling variability rather than memorization. DS-R1 Distill Llama 8B shows a larger gap (53.2\% vs.\ 63.2\%), but given its high item-level variance (53--70\% across prompt conditions), this likely reflects measurement noise rather than contamination. Overall, the close agreement between a dataset the models could have seen and one they could not provides reasonable evidence that the results reported here reflect genuine reasoning rather than recall of memorized answers. Finally, the evaluation is restricted to the multiple-choice format of the LSAT, and performance on constructed-response legal reasoning tasks may differ substantially.

The process supervision pilot itself is the most acknowledged of these limitations, having been conducted on a single model at a single scale with only 209 training examples. The resulting self-PRM improves DeepSeek-R1 Distill 7B from 51.9\% to 63.6\% on the held-out Official Test, a ceiling that remains far from frontier performance. Scaling the candidate pool (Best-of-$N$ with larger $N$), expanding the PRM training set through synthetic explanation generation, and applying process supervision during fine-tuning rather than only at inference time are all natural next steps within the small-model regime.

An equally compelling direction is applying process supervision to frontier models themselves. The small-model experiments were constrained to the scale at which fine-tuning remains feasible on a single GPU. Frontier models, which already score 97--99\%, make only a handful of errors, but those errors are deterministic and resistant to self-consistency. If a self-PRM can identify superior thinking traces among 7B-scale candidates, the same approach applied to frontier-scale candidates, where the base reasoning rigor is far higher, may push accuracy closer to the theoretical ceiling.

\section{Conclusion}

For the first time, a reasoning model has answered every question correctly on a test that the legal industry has spent three-quarters of a century using to identify its most capable minds.

Decades of correlation studies identify LSAT scores as the strongest single predictor of first-year law-school grades, a finding replicated as recently as the 2020--2024 admission cycles \citep{stilwell2011predictive, topczewski2025lsat}. Independent analyses trace the same time-pressured analytic signal into the in-class examinations students sit years later in law school itself \citep{henderson2004lsat}. A perfect 180 on this test is therefore among the most rigorously validated markers of elite human analytic capacity.

That a reasoning model now attains the same score, however, removes the LSAT from its long-standing position as a threshold only the human mind could cross. This does not diminish the reasoning skills the LSAT measures, nor does it render the exam irrelevant for admissions. But it does mean that the ceiling the LSAT was designed to measure, the upper bound of logical reasoning ability, is no longer exclusive to human cognition. For law schools and the legal profession, this should prompt a reckoning with what remains distinctively human about the capacity their gatekeeping exam was designed to measure, and with the possibility that this capacity has ceased to be the exclusive preserve of the human mind.

\bibliographystyle{unsrtnat}
\bibliography{references}

\addtocontents{toc}{\protect\setcounter{tocdepth}{2}}
\clearpage
\appendix

\section*{Appendix}
\makeatletter\@starttoc{toc}\makeatother
\clearpage

\newtcolorbox{appqbox}{
  enhanced,
  colback=white, colframe=gray!55, boxrule=0.5pt, arc=2pt,
  left=8pt, right=8pt, top=7pt, bottom=7pt,
  fontupper=\footnotesize,
  halign=flush left,
  title={\scriptsize\bfseries LSAT Official Test $\cdot$ April 2025 $\cdot$ Logical Reasoning $\cdot$ Question 5},
  colbacktitle=gray!18, coltitle=black,
  fonttitle=\scriptsize\bfseries,
  attach boxed title to top left={yshift=-2pt, xshift=8pt},
  boxed title style={colback=gray!18, colframe=gray!55, boxrule=0.4pt, arc=1pt,
                     left=4pt, right=4pt, top=1.5pt, bottom=1.5pt}
}
\newtcolorbox{athinkbox}[1][]{
  enhanced, breakable, enhanced jigsaw,
  colback=green!2, colframe=green!50!black, boxrule=0.5pt, arc=2pt,
  left=6pt, right=6pt, top=8pt, bottom=5pt,
  fontupper=\scriptsize,
  title={\scriptsize\bfseries Thinking Trace#1},
  colbacktitle=gray!18, coltitle=black,
  fonttitle=\scriptsize\bfseries,
  attach boxed title to top left={yshift=-2pt, xshift=8pt},
  boxed title style={colback=gray!18, colframe=gray!55, boxrule=0.4pt, arc=1pt,
                     left=4pt, right=4pt, top=1.5pt, bottom=1.5pt},
  before skip=2pt, after skip=2pt
}
\newtcolorbox{athinkboxweak}[1][]{
  enhanced, breakable, enhanced jigsaw,
  colback=red!2, colframe=red!50!black, boxrule=0.5pt, arc=2pt,
  left=6pt, right=6pt, top=8pt, bottom=5pt,
  fontupper=\scriptsize,
  title={\scriptsize\bfseries Thinking Trace#1},
  colbacktitle=gray!18, coltitle=black,
  fonttitle=\scriptsize\bfseries,
  attach boxed title to top left={yshift=-2pt, xshift=8pt},
  boxed title style={colback=gray!18, colframe=gray!55, boxrule=0.4pt, arc=1pt,
                     left=4pt, right=4pt, top=1.5pt, bottom=1.5pt},
  before skip=2pt, after skip=2pt
}
\newtcolorbox{arespbox}[1][]{
  enhanced, breakable, enhanced jigsaw,
  colback=green!2, colframe=green!50!black, boxrule=0.5pt, arc=2pt,
  left=6pt, right=6pt, top=8pt, bottom=5pt,
  fontupper=\scriptsize,
  title={\scriptsize\bfseries Response Trace#1},
  colbacktitle=gray!18, coltitle=black,
  fonttitle=\scriptsize\bfseries,
  attach boxed title to top left={yshift=-2pt, xshift=8pt},
  boxed title style={colback=gray!18, colframe=gray!55, boxrule=0.4pt, arc=1pt,
                     left=4pt, right=4pt, top=1.5pt, bottom=1.5pt},
  before skip=2pt, after skip=2pt
}
\newtcolorbox{arespboxweak}[1][]{
  enhanced, breakable, enhanced jigsaw,
  colback=red!2, colframe=red!50!black, boxrule=0.5pt, arc=2pt,
  left=6pt, right=6pt, top=8pt, bottom=5pt,
  fontupper=\scriptsize,
  title={\scriptsize\bfseries Response Trace#1},
  colbacktitle=gray!18, coltitle=black,
  fonttitle=\scriptsize\bfseries,
  attach boxed title to top left={yshift=-2pt, xshift=8pt},
  boxed title style={colback=gray!18, colframe=gray!55, boxrule=0.4pt, arc=1pt,
                     left=4pt, right=4pt, top=1.5pt, bottom=1.5pt},
  before skip=2pt, after skip=2pt
}

\newcommand{\correct}{{\color{green!55!black}\checkmark}}
\newcommand{\wrong}{{\color{red!55!black}$\boldsymbol{\times}$}}
\newcommand{\modelhdr}[2]{%
  \refstepcounter{subsection}%
  \addcontentsline{toc}{subsection}{\protect\numberline{\thesubsection}#1}%
  \par\bigskip\bigskip\noindent
  {\large\bfseries #1}\;{\small\color{gray!55!black}#2}\par
  \vspace{4pt}
}

\section{Full Model Thinking Traces and Responses on a Logical Reasoning Question from the Official Test}
\label{app:traces}

This appendix presents the complete thinking traces and responses of all eight evaluated models on the Logical Reasoning question abbreviated in Figure~\ref{fig:formquality}, alongside the official LSAC question stem and all five answer choices. All six frontier models answer correctly; both distilled small models select incorrect answers.

\begin{appqbox}
Researcher: Some scholars maintain that the first government other than China to issue any form of paper currency was the Massachusetts colony in 1690. But these scholars are incorrect. During a coin shortage in New France in 1685, its government was unable to use coins to pay soldiers stationed in Quebec. Instead, playing cards were assigned an official value and issued as pay. These cards, which could be used to buy goods, were convertible to coins, which later arrived on supply ships from France.\\[5pt]
\textbf{The researcher's conclusion is properly drawn if which one of the following is assumed?}
\begin{itemize}[noitemsep, topsep=3pt, leftmargin=1.6em, label={}]
  \item (A) The playing cards in use in Quebec in the late 1600s were made of paper. \hfill {\color{green!55!black}\textbf{[correct]}}
  \item (B) The Chinese government did not issue paper currency during the 1600s.
  \item (C) Playing cards were not available to inhabitants of the Massachusetts colony before 1690.
  \item (D) The Massachusetts colony's issue of paper currency was not prompted by a coin shortage.
  \item (E) Every government that issued paper currency in the 1600s left records pertaining to that currency.
\end{itemize}
\end{appqbox}

\modelhdr{GPT-5}{(OpenAI)}
\begin{athinkbox}[ $\cdot$ 3{,}428 chars]
\textbf{Analyzing LSAT question}\par\smallskip
I need to tackle an LSAT LR question, specifically looking for what assumption makes the researcher's conclusion valid. The stimulus presents a claim that the Massachusetts colony was the first to issue paper currency, but argues this is incorrect due to an earlier instance in New France with playing cards. I should identify the assumption that connects these points logically to support the conclusion. So, I'm gathering details carefully to solve it.

\textbf{Identifying the assumption}\par\smallskip
I'm analyzing a conclusion that says scholars are wrong about the Massachusetts colony being the first to issue paper currency. The evidence points to New France issuing playing cards as payment in 1685, which were convertible to coins. To support the conclusion, I need an assumption that those playing cards qualify as ``paper currency.'' If they don't fit that definition, the evidence doesn't effectively counter the scholars' claim. So now, I should zero in on which option aligns with that requirement.

\textbf{Exploring the options}\par\smallskip
Option (A) contends that the playing cards in Quebec during the late 1600s were made of paper. If that's true, and these cards had official value as payment and could be converted to coins, they would resemble currency. However, the key question is whether they qualify as ``paper currency.'' They need to be both paper and currency. While playing cards are often made from paper, LSAT questions require precision about that aspect. So, I'll check Option (B), which states that the Chinese government didn't issue paper currency in the 1600s.

\textbf{Reviewing response options}\par\smallskip
The scholars claim Massachusetts was the first to issue any paper currency besides China in 1690, but the researcher argues they are wrong because New France did it earlier in 1685. The status of China's actions isn't essential to the argument.

Options (C) and (D) don't provide necessary information, and (E) is also irrelevant since the argument only relies on one example from New France. Option (A) looks most reasonable, but I need to double-check if playing cards truly count as ``paper.''

\textbf{Finalizing assumptions for LSAT}\par\smallskip
I'm exploring whether the playing cards issued in New France qualify as ``paper currency.'' They were assigned official value, used as pay, and could be converted to coins, suggesting they acted as currency. Option (A) asserts that the cards were made of paper, which connects them to the concept of ``paper currency.'' Other options, like (B), (C), (D), and (E), don't offer relevant information. The crux is that while the cards function as currency, confirming they're made of paper is crucial. Thus, (A) is necessary for the conclusion to hold.

\textbf{Finalizing LSAT answer}\par\smallskip
I need to make sure to assume the playing cards were made of paper, which leads me to choose option (A). Many test-takers might be quick to pick this answer, but I'll craft a clear response: the correct answer is (A). The researcher refutes the scholars' claim by citing an earlier instance of paper currency in New France. Therefore, with the assumption that the cards were made of paper, it shows that the scholars are incorrect regarding Massachusetts in 1690.

Answer: A. Explanation: The researcher indicates that New France issued currency-like instruments in 1685; to establish they were paper currency, we assume the playing cards were made of paper, refuting the claim about Massachusetts.
\end{athinkbox}

\begin{arespbox}[ $\cdot$ Answer: (A) \correct]
A

Explanation: The argument refutes the scholars by giving an earlier instance of ``paper currency'' in New France. The evidence shows the cards functioned as currency; to make them ``paper currency,'' we must assume the playing cards were made of paper.
\end{arespbox}

\modelhdr{Claude Opus 4}{(Anthropic)}
\begin{athinkbox}[ $\cdot$ 2{,}207 chars]
Let me break down this argument:

The researcher's position:
\begin{itemize}[noitemsep, topsep=2pt, leftmargin=1.4em]
\item Some scholars say Massachusetts was the first government other than China to issue paper currency (in 1690)
\item The researcher says these scholars are incorrect
\item The researcher's evidence: New France issued playing cards as currency in 1685 (5 years before Massachusetts)
\end{itemize}

For the researcher's conclusion to be ``properly drawn,'' we need to figure out what assumption would make this argument valid.

The researcher is essentially saying: ``Massachusetts wasn't the first non-Chinese government to issue paper currency because New France did it earlier with playing cards.''

For this to work as a counterargument, the playing cards must qualify as ``paper currency.''

Let me examine each answer choice:

(A) The playing cards in use in Quebec in the late 1600s were made of paper.
\begin{itemize}[noitemsep, topsep=2pt, leftmargin=1.4em]
\item This would be necessary for the playing cards to count as ``paper currency''
\item If they weren't made of paper, then they wouldn't be paper currency, and the researcher's counterexample wouldn't work
\end{itemize}

(B) The Chinese government did not issue paper currency during the 1600s.
\begin{itemize}[noitemsep, topsep=2pt, leftmargin=1.4em]
\item This is irrelevant to the argument
\item The claim is about the first government ``other than China'' --- it doesn't matter when China issued paper currency
\end{itemize}

(C) Playing cards were not available to inhabitants of the Massachusetts colony before 1690.
\begin{itemize}[noitemsep, topsep=2pt, leftmargin=1.4em]
\item This is irrelevant
\item The argument is about New France using playing cards, not Massachusetts
\end{itemize}

(D) The Massachusetts colony's issue of paper currency was not prompted by a coin shortage.
\begin{itemize}[noitemsep, topsep=2pt, leftmargin=1.4em]
\item This is irrelevant to whether Massachusetts was first
\item The reason for issuing doesn't affect the timing
\end{itemize}

(E) Every government that issued paper currency in the 1600s left records pertaining to that currency.
\begin{itemize}[noitemsep, topsep=2pt, leftmargin=1.4em]
\item This might help establish historical facts but isn't necessary for this specific argument
\item The researcher already has evidence about New France
\end{itemize}

The key insight is that the researcher is using playing cards as an example of ``paper currency'' to show that New France issued it before Massachusetts. For this argument to work, the playing cards must actually be made of paper --- otherwise they wouldn't be ``paper currency'' at all.

The answer is (A).
\end{athinkbox}

\begin{arespbox}[ $\cdot$ Answer: (A) \correct]
Looking at this argument, I need to identify what assumption would make the researcher's conclusion valid.

The researcher's argument structure:
\begin{itemize}[noitemsep, topsep=2pt, leftmargin=1.4em]
\item Scholars claim: Massachusetts was the first government (other than China) to issue paper currency in 1690
\item Researcher's counterargument: New France issued playing cards as currency in 1685, which is earlier than Massachusetts
\item Therefore: The scholars are incorrect
\end{itemize}

For this counterargument to work, the playing cards from New France must qualify as ``paper currency.'' The researcher is using the 1685 playing cards as evidence that someone issued paper currency before Massachusetts did in 1690.

Let me evaluate each choice:

(A) The playing cards in use in Quebec in the late 1600s were made of paper.
\begin{itemize}[noitemsep, topsep=2pt, leftmargin=1.4em]
\item This is crucial! If the playing cards weren't made of paper, they wouldn't be ``paper currency,'' and the researcher's counterexample would fail.
\end{itemize}

(B) The Chinese government did not issue paper currency during the 1600s.
\begin{itemize}[noitemsep, topsep=2pt, leftmargin=1.4em]
\item Irrelevant --- the claim explicitly says ``other than China''
\end{itemize}

(C) Playing cards were not available to inhabitants of the Massachusetts colony before 1690.
\begin{itemize}[noitemsep, topsep=2pt, leftmargin=1.4em]
\item Irrelevant --- this is about New France's actions, not Massachusetts'
\end{itemize}

(D) The Massachusetts colony's issue of paper currency was not prompted by a coin shortage.
\begin{itemize}[noitemsep, topsep=2pt, leftmargin=1.4em]
\item Irrelevant --- the motivation doesn't affect whether they were first
\end{itemize}

(E) Every government that issued paper currency in the 1600s left records pertaining to that currency.
\begin{itemize}[noitemsep, topsep=2pt, leftmargin=1.4em]
\item Too broad and unnecessary for this specific argument
\end{itemize}

The answer is \textbf{(A)}. The researcher's conclusion depends on the playing cards being made of paper so they can count as ``paper currency.''
\end{arespbox}

\modelhdr{Gemini 2.5 Pro}{(Google)}
\begin{athinkbox}[ $\cdot$ 2{,}791 chars]
\textbf{Deconstructing the Argument: The Necessary Assumption}\par\smallskip
Okay, so I'm presented with a Logical Reasoning question. My primary task is to find the necessary assumption, that critical unstated link that the argument \textit{needs} to work. I begin by dissecting the argument: the scholars are saying Massachusetts was the first, other than China, to issue paper currency. The author claims they're wrong, pointing to a New France example from 1685. These playing cards, used during a coin shortage, functioned like currency.

Immediately, I see a potential weak point: the scholars are discussing \textit{paper currency}, while the argument presents playing cards. The leap from ``playing cards used as currency'' to ``paper currency'' is the crux of this question. The argument hinges on those playing cards \textit{being} paper currency.

My analysis zeroes in on this gap. The text \textit{describes} the function of these cards --- official value, used for purchases, convertible to coins --- but it's conspicuously silent on \textit{what they were made of}. If they were made of something other than paper, like wood or leather, the argument crumbles. The conclusion collapses without that critical assumption.

Now, I look at the answer choices. I need an assumption that connects the New France example to the \textit{paper} part of the scholars' claim.

\textbf{(A) The playing cards in use in Quebec in the late 1600s were made of paper.} Bingo! This fills the gap perfectly. If the cards were paper, the evidence directly refutes the scholars' claim. I can even apply the Negation Test to confirm: If they were \textit{not} made of paper, the argument collapses, because the example is no longer relevant. This is a very strong contender.

I quickly check the remaining options to ensure that they are irrelevant:
\begin{itemize}[noitemsep, topsep=2pt, leftmargin=1.4em]
\item \textbf{(B) The Chinese government did not issue paper currency during the 1600s.} Irrelevant. The question is who issued the \textit{first} paper currency \textit{other than China}.
\item \textbf{(C) Playing cards were not available to inhabitants of the Massachusetts colony before 1690.} Irrelevant. The argument focuses on New France, not Massachusetts.
\item \textbf{(D) The Massachusetts colony's issue of paper currency was not prompted by a coin shortage.} Irrelevant. Why Massachusetts issued the currency is immaterial. The argument hinges on the fact that the New France \textit{example} happened earlier.
\item \textbf{(E) Every government that issued paper currency in the 1600s left records pertaining to that currency.} Irrelevant. The argument relies on specific examples, and it does not need to consider a \textit{universal} rule.
\end{itemize}

I return to option (A) and it is the necessary assumption. Without that assumption, the argument is unsound. The playing cards must be \textit{made} of paper for the argument to succeed. My analysis concludes, and I select option (A).
\end{athinkbox}

\begin{arespbox}[ $\cdot$ Answer: (A) \correct]
This is an excellent example of an LSAT ``Assumption'' question. Here's a breakdown of the argument and the correct answer:

\textbf{Deconstructing the Argument}\par\smallskip
\begin{enumerate}[noitemsep, topsep=2pt, leftmargin=1.6em]
\item \textbf{Scholars' Claim:} The first government (besides China) to issue \textit{paper currency} was Massachusetts in 1690.
\item \textbf{Researcher's Conclusion:} The scholars are incorrect.
\item \textbf{Researcher's Evidence:} In New France in 1685 (which is 5 years \textit{before} 1690), the government issued playing cards as currency. These cards had official value, could be used for purchases, and were redeemable for coins.
\end{enumerate}

\textbf{Identifying the Logical Gap}\par\smallskip
The researcher is trying to prove the scholars wrong about \textit{paper currency}. To do this, the researcher provides an example of \textit{playing cards} being used as currency. The entire argument rests on the idea that this example of playing cards \textit{is} an example of paper currency.

However, the argument never explicitly states what the playing cards were made of. For the researcher's evidence to actually contradict the scholars' claim, the playing cards from 1685 must qualify as ``paper currency.'' This means they must have been made of paper. This is the unstated assumption that links the evidence to the conclusion.

\textbf{Analyzing the Answer Choices}\par\smallskip
\begin{itemize}[noitemsep, topsep=2pt, leftmargin=1.4em]
\item \textbf{(A) The playing cards in use in Quebec in the late 1600s were made of paper.} This directly fills the logical gap identified above. If the playing cards were made of paper, then the New France government did indeed issue \textit{paper currency} in 1685, making the scholars' claim (that Massachusetts was first in 1690) incorrect. This is the necessary assumption.
\item \textbf{(B) The Chinese government did not issue paper currency during the 1600s.} This is irrelevant. The scholars' claim is about the first government \textit{other than China}. What China did during that time doesn't affect the dispute about who came first between New France and Massachusetts.
\item \textbf{(C) Playing cards were not available to inhabitants of the Massachusetts colony before 1690.} The availability of playing cards to the general public in Massachusetts is irrelevant. The argument is about what the \textit{government} issued as currency.
\item \textbf{(D) The Massachusetts colony's issue of paper currency was not prompted by a coin shortage.} The \textit{reason} for issuing currency (whether a coin shortage or something else) is not relevant to the core of the argument, which is about \textit{when} it was issued and \textit{what} it was made of.
\item \textbf{(E) Every government that issued paper currency in the 1600s left records pertaining to that currency.} This is too broad. The researcher's argument only relies on the existence of records for the specific cases of New France and Massachusetts. It doesn't need to assume that records exist for \textit{every} case.
\end{itemize}

\textbf{Conclusion}\par\smallskip
The researcher's argument only works if the playing cards used as currency in New France are considered an example of ``paper currency.'' For that to be true, they must have been made of paper.

Therefore, the correct answer is \textbf{(A)}.
\end{arespbox}

\modelhdr{DeepSeek-R1}{(DeepSeek, 671B)}
\begin{athinkbox}[ $\cdot$ 5{,}202 chars]
We are given an LSAT Logical Reasoning question. The researcher argues that some scholars maintain that the first government other than China to issue any form of paper currency was the Massachusetts colony in 1690. But the researcher says these scholars are incorrect. The researcher provides evidence: During a coin shortage in New France in 1685, its government was unable to use coins to pay soldiers stationed in Quebec. Instead, playing cards were assigned an official value and issued as pay. These cards, which could be used to buy goods, were convertible to coins, which later arrived on supply ships from France.

The researcher's conclusion is that the scholars are incorrect (i.e., Massachusetts colony was not the first government other than China to issue paper currency). The evidence suggests that New France in 1685 issued playing cards as a form of currency. So the researcher is implicitly claiming that the playing cards issued by New France in 1685 constitute a form of paper currency, and since 1685 is earlier than 1690, New France, not Massachusetts, was the first (excluding China). But for this argument to work, we need to assume that the playing cards were indeed a form of paper currency. The question asks: ``The researcher's conclusion is properly drawn if which one of the following is assumed?'' This is a necessary assumption question. We need to find an assumption that is required for the conclusion to follow logically from the premises.

Let's analyze the argument structure:

Premise: In 1685, New France government issued playing cards as pay, assigned official value, convertible to coins, usable to buy goods.

Conclusion: Scholars are incorrect that Massachusetts colony in 1690 was the first government other than China to issue any form of paper currency.

The conclusion says ``paper currency.'' The evidence is about ``playing cards.'' For the conclusion to hold, we must assume that these playing cards count as paper currency. Otherwise, if they are not paper currency, then the fact that New France issued them doesn't challenge the scholars' claim about paper currency.

Option (A) says: ``The playing cards in use in Quebec in the late 1600s were made of paper.'' This directly connects playing cards to paper. If the playing cards were not made of paper, then they might not be considered ``paper currency.'' So this assumption seems necessary: if the playing cards were not made of paper, then they are not paper currency, and the argument fails. So (A) looks like a necessary assumption.

Check the other options:

(B) ``The Chinese government did not issue paper currency during the 1600s.'' The scholars' claim is about ``the first government other than China to issue any form of paper currency.'' That implies that China issued paper currency earlier, but they are excluding China. The researcher is not challenging that China had paper currency earlier. The researcher is only saying that Massachusetts was not the first non-China government; New France was earlier. For the argument, we don't need to assume anything about when China issued paper currency. The scholars already exclude China, so even if China issued paper currency in the 1600s, that doesn't affect the comparison between New France and Massachusetts. Actually, the scholars likely think China issued paper currency much earlier. So (B) is not necessary.

(C) ``Playing cards were not available to inhabitants of the Massachusetts colony before 1690.'' This is irrelevant. Whether playing cards were available in Massachusetts doesn't affect whether New France issued paper currency before Massachusetts. The issue is about issuance by governments, not availability.

(D) ``The Massachusetts colony's issue of paper currency was not prompted by a coin shortage.'' This is irrelevant. The reason for Massachusetts issuing paper currency doesn't affect whether New France issued paper currency earlier. The conclusion is about which was first, not why they issued it.

(E) ``Every government that issued paper currency in the 1600s left records pertaining to that currency.'' This is about records. The argument doesn't depend on whether records were left. The evidence about New France comes from historical accounts, so presumably there are records. But even if not all governments left records, it doesn't undermine the specific evidence about New France. This is not necessary for the conclusion.

Thus, (A) is the most clearly necessary assumption. Without (A), if the playing cards were not made of paper, then they are not paper currency, so New France did not issue paper currency, and the scholars might still be correct. So the argument requires that the playing cards were made of paper.

Confirm: The conclusion says ``paper currency.'' The playing cards are described as being assigned official value, used to buy goods, convertible to coins. That sounds like currency. But if they are not made of paper, they are not paper currency. They could be made of cardboard or some other material. But typically, playing cards in that era were made of paper or cardstock. Still, we need to assume they were paper to call them paper currency. So (A) is necessary.

Therefore, the answer is (A).
\end{athinkbox}

\begin{arespbox}[ $\cdot$ Answer: (A) \correct]
The argument concludes that the scholars are incorrect in claiming that the Massachusetts colony in 1690 was the first government other than China to issue paper currency, citing the earlier use of playing cards as currency in New France in 1685. For this conclusion to hold, the playing cards must qualify as paper currency.

\begin{itemize}[noitemsep, topsep=2pt, leftmargin=1.4em]
\item \textbf{Option (A)} is necessary because if the playing cards were not made of paper, they would not constitute ``paper currency,'' undermining the reasoning that New France issued paper currency before Massachusetts.
\item \textbf{Option (B)} is irrelevant; the scholars already exclude China, so whether China issued paper currency in the 1600s does not affect the comparison between New France and Massachusetts.
\item \textbf{Option (C)} is irrelevant; the availability of playing cards in Massachusetts has no bearing on which government first issued paper currency.
\item \textbf{Option (D)} is irrelevant; the motivation behind Massachusetts's issuance does not impact the chronological order.
\item \textbf{Option (E)} is unnecessary; the existence of records for all issuers is not required to support the specific evidence about New France.
\end{itemize}

Thus, the argument assumes that the playing cards were made of paper, making (A) the correct answer.

\textbf{ANSWER: (A)}
\end{arespbox}

\modelhdr{Kimi K2 Thinking}{(Moonshot AI, 1T)}
\begin{athinkbox}[ $\cdot$ 4{,}851 chars]
We need to parse the LSAT logical reasoning question. The researcher says: Some scholars maintain that the first government other than China to issue any form of paper currency was the Massachusetts colony in 1690. But these scholars are incorrect. The researcher then provides an example: During a coin shortage in New France in 1685, its government was unable to use coins to pay soldiers stationed in Quebec. Instead, playing cards were assigned an official value and issued as pay. These cards, which could be used to buy goods, were convertible to coins, which later arrived on supply ships from France.

We need to find an assumption that, if true, makes the conclusion properly drawn. The conclusion is that the scholars are incorrect: the first government other than China to issue any form of paper currency was not Massachusetts in 1690; rather, New France in 1685 issued playing cards as currency, which qualifies as paper currency.

Thus, the argument is: New France's playing cards were a form of paper currency, and they were issued in 1685, earlier than 1690. So the scholars are wrong.

We need an assumption that bridges the gap between the premise (cards were assigned official value, used as pay, convertible to coins) and the conclusion (this qualifies as ``any form of paper currency''). The key is that the playing cards must be considered ``paper currency.'' The researcher says ``playing cards were assigned an official value and issued as pay. These cards, which could be used to buy goods, were convertible to coins.'' But is that enough to be ``paper currency''? The assumption must be that playing cards are made of paper. That would be the missing link: If the cards are made of paper, then they are a form of paper currency. So answer (A) seems likely.

But we need to examine each answer choice:

(A) The playing cards in use in Quebec in the late 1600s were made of paper.

If this is true, then the playing cards are indeed paper currency. The researcher didn't explicitly state that they were made of paper, but the conclusion is about paper currency. So this is a necessary assumption: that the cards are paper. Without that, the argument fails. So (A) is a strong candidate.

(B) The Chinese government did not issue paper currency during the 1600s.

This is irrelevant to the conclusion; the conclusion is about the first government other than China. Chinese issuance is not relevant.

(C) Playing cards were not available to inhabitants of the Massachusetts colony before 1690.

This is about availability, but not about the definition of paper currency.

(D) The Massachusetts colony's issue of paper currency was not prompted by a coin shortage.

Irrelevant.

(E) Every government that issued paper currency in the 1600s left records pertaining to that currency.

Irrelevant.

Thus, answer is (A). However, we need to check if any other answer might be necessary. Let's think: The conclusion is ``these scholars are incorrect'' because the first government other than China to issue any form of paper currency was not Massachusetts in 1690; it was New France in 1685. The evidence: New France issued playing cards as pay, convertible to coins. The missing premise is that playing cards count as paper currency. The assumption could be that playing cards are made of paper. Or perhaps that ``any form of paper currency'' includes playing cards that are assigned an official value and convertible to coins. But does ``paper currency'' require that the instrument be made of paper? Usually, yes. If the cards were made of something else (e.g., ivory, wood), they might not be considered ``paper currency.'' So the assumption that they are made of paper is needed.

But maybe the assumption is something else: The researcher says ``these cards, which could be used to buy goods, were convertible to coins.'' That might be enough to consider them as currency. But the phrase ``paper currency'' also requires that they be made of paper. So (A) is necessary.

Alternatively, maybe the assumption is that the playing cards were ``issued by the government.'' But the researcher says ``its government was unable to use coins to pay soldiers\ldots{} Instead, playing cards were assigned an official value and issued as pay.'' So it's clear that the government assigned value and issued them. So that's not missing.

Thus, answer (A) is correct.

We need to verify that (A) is indeed an assumption, not a stated premise. The researcher never says they were made of paper; they say ``playing cards.'' Usually playing cards are made of paper (or cardstock). But historically, playing cards could be made of other materials? In the 1600s, playing cards were typically made of paper or cardstock. But the argument doesn't explicitly state that they were paper. So the assumption is that they were made of paper.

Thus answer: (A).

Thus, the answer is (A).
\end{athinkbox}

\begin{arespbox}[ $\cdot$ Answer: (A) \correct]
\textbf{Answer: (A) The playing cards in use in Quebec in the late 1600s were made of paper.}

\textbf{Explanation}\par\smallskip
The researcher's argument is:
\begin{enumerate}[noitemsep, topsep=2pt, leftmargin=1.6em]
\item Some scholars claim that Massachusetts (1690) was the first government other than China to issue paper currency.
\item In 1685 New France paid soldiers with playing cards that had an official value, could be used to buy goods, and were convertible to coins.
\item Therefore, the scholars are wrong --- New France's playing cards predate Massachusetts's paper currency.
\end{enumerate}

The gap between premise (2) and the conclusion is that the playing cards must actually count as \textit{paper} currency. The researcher never states what material the cards were made of; if they were, say, ivory or wood, they wouldn't be ``paper currency.'' Hence the argument depends on the unstated assumption that those cards were indeed made of paper. Only (A) supplies this necessary link.

The other choices are irrelevant to the claim about ``paper currency'' and do not affect the logical connection between the evidence and the conclusion.
\end{arespbox}

\modelhdr{QwQ-32B}{(Alibaba/Qwen, 32B)}
\begin{athinkbox}[ $\cdot$ 16{,}780 chars]
Okay, so I need to figure out which assumption is necessary for the researcher's conclusion to be properly drawn. Let's start by understanding the argument here.

The researcher is arguing that some scholars are incorrect in saying that the first government other than China to issue paper currency was Massachusetts in 1690. The researcher's counterexample is New France in 1685, where they used playing cards as currency during a coin shortage. The cards had official value, could be used to buy goods, and were convertible to coins later.

So the conclusion is that Massachusetts wasn't the first; New France was. But for that to hold, what needs to be assumed? The question is asking for the necessary assumption here, which is an unstated premise required for the conclusion to follow logically.

Let me look at the options.

Option A says the playing cards were made of paper. Hmm, because the researcher refers to them as "paper currency," right? Because if the playing cards weren't made of paper, then maybe they wouldn't count as paper currency. Like, if they were made of some other material, maybe the scholars' definition of paper currency would exclude them. So A might be a necessary assumption. The researcher needs to assume that the playing cards were actually made of paper to classify them as paper currency. That seems possible.

Option B says the Chinese government didn't issue paper currency in the 1600s. Wait, the original claim is that the first after China was Massachusetts. The researcher's point is that before 1690, another one (New France) was earlier. But the scholars say "other than China," so maybe they already accounted for China? Wait, no, the claim is that the first government \textit{other than China} was Massachusetts in 1690. So the researcher's counter is that New France was first other than China. Therefore, the researcher is not challenging the scholars on whether China had it earlier; in fact, the scholars probably already accept that China was first, but the researcher is saying that the next one, the first non-China one, was New France in 1685 instead of Mass in 1690. Therefore, does the researcher need to assume that the Chinese government didn't do paper currency in 1685 or 1690? Wait, but the conclusion is only that the first non-China was New France, not that China didn't have it earlier. Maybe the scholars already agree China was first, so the researcher's argument doesn't need to assume that China didn't also have some later currency. Wait, but the original statement by the scholars might have been that the first government other than China was Massachusetts. For that to be the case, they must believe that China was earlier. However, maybe the researcher's point is that China might have had it even later, but that might not be relevant here because the question is about non-China. Hmm. Let me think again.

If the researcher's argument is that New France's 1685 was the first non-China, then to make that conclusion, they have to assume that no other non-China government had it before 1685. Wait, but how does that tie into the options here? Option B is about China's issuing in the 1600s. So if the researcher's argument requires that China did not have any paper currency in the 1600s, then that would be needed, but perhaps not. Wait, no. Because the original scholars say that the first after China was Mass 1690. To refute that, the researcher only needs to establish that another non-China government (New France) did it earlier (1685). Therefore, the researcher doesn't need to assume that China itself didn't do anything in the 1600s, because even if China did have it in, say, 1600, then the first after China would be the next, and here is 1685 New France. Therefore, the B option saying that China didn't issue in the 1600s is not necessary for the conclusion? Because the existence or non-existence of Chinese paper currency in the 1600s is separate from the issue between Mass and New France. Because the comparison is between non-China, so even if China did continue to issue paper currency throughout the 1600s, as long as New France did before 1690, then that's still valid as the "first non-China." Wait, but actually, the scholars state that Mass was the first non-China, so to reject that, the researcher must show someone else was first non-China. So maybe B is not needed. Because even if China had more recent issuances, like in 1600, it doesn't affect that New France in 1685 would still be the first post-China? Wait maybe the problem is that maybe the scholars think that China had paper currency but in the 1500s, so the next one (other than China) is Mass in 1690, but the researcher is saying New France in 1685 was earlier than Mass and also not China, so that's valid. Therefore, the researcher does not have to assume that China didn't have paper currency in the 1600s. Because even if China had it in the 1600s, the first non-China is still New France. Wait, actually that might not make sense. Suppose China had issued in 1680, then the ``first government other than China'' would require going after China's issuance. Wait maybe I'm overcomplicating.

Let me think again of the options. Option B: The Chinese government did not issue paper currency during the 1600s. Wait, when scholars say "the first government other than China", they must be assuming that China did have paper currency before any other governments. Therefore, to accept that Massachusetts was first after China, they would have had to accept that China had paper currency prior to that. However, the researcher is pointing out New France in 1685 as the first (so arguing against Mass in 1690 as first after China). The argument would require that China wasn't also doing it in 1685. Wait, actually no. Suppose that in 1680 China had issued paper and then in 1685, China also started doing something else. But the claim is that the first "other than China" is the first non-China one. So regardless of what China did after that, as long as New France was non-China and earlier than Massachusetts, then New France would be the first after China. So the Chinese activities in the 1600s after their initial issuing wouldn't affect the fact that New France's issuance was before Mass.

Therefore, the researcher could still be right even if China had some currency in the 1600s. Therefore, B isn't required. So maybe B is not the assumption.

Option C: Playing cards weren't available in Massachusetts before 1690. Hmm, does the argument depend on that? The researcher is talking about New France, so the Massachusetts having or not having cards before 1690 might not be relevant. The conclusion is that MA wasn't the first because another entity did it earlier, so unless the MA people had done something even earlier with paper currency, which would challenge New France's example.

Wait but the argument is that New France beat Massachusetts by 5 years. The scholars say the first "other than China" is MA in 1690. The researcher counters with NF in 1685. To make that conclusion, the researcher has to ensure that the NF case counts as "paper currency" and that no other government (other than China) did it before 1685. The question is, for the conclusion to hold, what must be true?

Option C is about Massachusetts not having it before 1690. But maybe another place could have done it between 1685 and 1690 but also before Massachusetts? Wait no, the researchers aren't addressing that. The argument is specifically against Massachusetts being the first after China, so pointing out that NF was earlier than MA. The researcher does not need to assume Massachusetts didn't have paper currency before 1690, because their point is that even if MA started in 1690, the NF was earlier. However, maybe if Massachusetts had paper before 1690, that could complicate things but the conclusion is about being first after China, so if NF is before MA, and before 1690, then even if MA had done it in 1685 too, but actually the scholars said MA was first in 1690, so the researcher's proof is that NF was first in 1685, so Ma's own starts after that. Therefore, the argument doesn't require assuming that MA didn't have it before 1690. Unless the problem is that maybe the Massachusetts used paper money before 1690 but later the scholars think it was 1690. Hmm, but the question states that the scholar's maintain that the first other than China was in 1690, so the argument may not require that Massachusetts didn't start earlier, unless perhaps the playing cards in MA before 1690 would disprove the scholar? Wait, but the researcher isn't talking about Massachusetts. The researcher's only point is about NF being earlier than MA. So maybe C isn't necessary.

Option D says Massachusetts's issuance wasn't prompted by a coin shortage. The researcher mentioned that in New France the coin shortage was the reason. If someone says that MA did it motivated by a different reason, would that affect? Not sure. The reasoning here: the researcher is pointing out an earlier case, so MA's motivation (shortage or not) doesn't really impact the timing. So D might not be necessary.

Option E: Every government that issued paper currency in the 1600s left records. Hmm, so the researcher is using the records of the New France's case, but for the conclusion that there wasn't any earlier instance other than China not accounted for, E would be something like "there are no such records." Wait, but E says every government that issued paper currency left records. So if E is assumed, then if there were any other government before 1685 (not China) that did it, their records would exist. So since we don't have such records, we can safely conclude that New France is the first. Wait but the conclusion is just about New France being earlier than Massachusetts based on the given info. However, maybe the researcher needs to assume that there wasn't some other earlier non-China government that we don't know about. If there were another one but no records (so E isn't true), then the researcher couldn't know. But the researcher is taking their example as sufficient. So the assumption would be that there are no earlier non-China cases that are unrecorded, and E would ensure that if they did exist, we would have records. Since the researcher is assuming that New France is the first non-China, maybe they assume there are no omitted cases.

Wait, but the question is "The conclusion is properly drawn if which one is assumed." So to make the researcher's argument valid, we have to assume that there are no other governments that issued earlier than 1685 except China. But how is that connected to the options?

Option E might be relevant here. If it's assumed that every government that did so left records, then since there are no records of any such earlier to 1685, the researcher can conclude that there are none. But the problem is that the question is structured as the researcher is contradicting the scholars' claim that MA was first. The researcher provides an example of earlier (NF 1685). But to make the conclusion that it's the first non-China, must the researcher assume that there is no other non-China before 1685? To do that, maybe the researcher assumes that the records of such a government would exist (so since none are found), there's none. That's possible, but the option E says every such government did leave records. So if that's true, then if there was any before NF, it would have records. Since the researcher didn't mention any, they can safely take their example as the first. So E is necessary. However, the problem is that the conclusion is only about the MA scholars' claim. The researchers don't need to account for 1680, unless the scholars' claim was that Mass was first, so as long as there's nothing between 1685 and 1690, but actually, the problem might not need to go that far.

Alternatively, maybe the key is that the playing cards were paper, hence A. The problem is whether the playing cards count as "paper currency." The term "paper currency" could be defined as paper, so if the cards were made of something else, maybe card stock is paper? Probably, but the researcher is assuming they are made of paper. The problem might consider that a playing card is on paper, but maybe not necessarily. Card (e.g., made of wood or something) is a term that originally could refer to card material, maybe? Wait, in 1685, playing cards would have been made of paper. Maybe it's common knowledge. However, questions need the answer to be necessary. The problem is, does the argument assume that the playing cards were made of paper? Because if they were not paper (say, wood or metal) then their currency would not be "paper currency" even if they were used as such, so then the researcher's example wouldn't refute the scholars, because the scholars are talking about a government other than China issuing paper currency. Therefore, A is a necessary assumption.

Between A and E. Hmm.

Wait let's think again. The scholars say Massachusetts was the first after China. The researcher says "No, because New France did so in 1685 with playing cards."

But unless those playing cards are considered paper, their currency (even if similar) wouldn't be paper. So for the researcher's counterexample to count as paper currency, the cards must have been made of paper. So without A, the researcher hasn't shown that the New France case was paper currency, so their argument would fail.

Therefore A must be assumed.

E is about records. The researcher must assume that there is no other earlier case, but unless E says that, how can they be sure? Well, the researcher is presenting their own example and saying that it's earlier than the scholars' candidate. But maybe someone else did it earlier? But to make the conclusion that Massachusetts isn't first, the presence of just one prior non-China case is sufficient. So even if there was a third case before both MA and New France, that would not affect the argument, because the conclusion is that the scholars are wrong in stating that MA was the first. The existence of NF alone is sufficient to make them wrong, so even if there was something earlier than both, but not China, the scholars would still be wrong because the early NF example has already occurred before MA. Therefore, the conclusion only requires that New France's example is valid and not refuted. So whether there's something before NF would only weaken the researcher's point that NF was first, but the conclusion is only countering that MA is first. Hence E may not be required because the argument doesn't need to establish that NF is the absolute first non-China, just that it's earlier than MA.

Therefore, the key assumption is A. The playing cards were made of paper. If they are not, then New France's issuance isn't paper currency, so the scholar's claim about Massachusetts might still hold.

So I think A is the correct answer.

Wait, but maybe someone might say E is important. Let's see. The researcher's argument relies on their knowledge of New France's case. If someone else did it earlier but there are no records because E isn't true, then the researcher's conclusion could be wrong, but the question is asking for an assumption required for the conclusion to be properly drawn. That is, the conclusion is: the scholars are incorrect because New France did it earlier. For this to hold, must there be no other prior cases that aren't recorded? The conclusion can still hold even if there are others. The researcher doesn't need to know that there's nothing else. The scholars claim that the first (other than China) was MA. For the researcher to show that someone else was first, you just need to have one example before MA. The existence of a third earlier case would not negate that. Therefore, E isn't necessary.

Therefore, A is the answer.

Wait another way to check. Suppose A is not true. The playing cards were made of leather or something else. Then when the researcher talks about using playing cards, that's not paper currency, so the scholars would still be correct that the first non-China paper currency was MA in 1690. Therefore, the conclusion's validity depends on A, hence A must be assumed.

Yes, so A is necessary.

Option B: If China did do paper currency in the 1600s, does that affect the conclusion? Let's see. Suppose that China had issued paper currency in, say, 1600, then the next non-China would be New France 1685. The claim against the scholars (who said MA was first) would still hold. If China had done it in 1690, then that would conflict, but the scholars are talking about the first other than China. The researcher just has to have NF before MA. So B doesn't seem necessary.

Therefore, the correct answer is A.
\end{athinkbox}

\begin{arespbox}[ $\cdot$ Answer: (A) \correct]
The researcher's argument concludes that Massachusetts was not the first government other than China to issue paper currency, as New France did so earlier in 1685 using playing cards. For this conclusion to hold, two key assumptions are critical:

\begin{enumerate}[noitemsep, topsep=2pt, leftmargin=1.6em]
\item \textbf{The playing cards must qualify as ``paper currency.''} The term ``paper currency'' implies the medium is made of paper. If the playing cards were constructed from a different material (e.g., wood, metal, or leather), the researcher's counterexample would not refute the scholars' claim about Massachusetts issuing the first non-China paper currency. Thus, \textbf{Option A} (``The playing cards in use in Quebec in the late 1600s were made of paper'') is a necessary assumption. Without it, the New France example loses its relevance to the definition of ``paper currency.''
\item \textbf{Other factors like availability of playing cards in Massachusetts (Option C)} or motivations (Option D) do not affect the logical link between New France's earlier issuance and the conclusion. Similarly, \textbf{Option B} (China not issuing currency in the 1600s) is unnecessary because the argument focuses on New France's \textit{earlier} issuance compared to Massachusetts, not directly on China's timeline. Even if China issued currency in the 1600s, New France would still qualify as the first non-China issuer after China.
\end{enumerate}

\textbf{Option E} (``Every government that issued paper currency in the 1600s left records'') is also not required. The researcher merely needs a valid example (New France in 1685) to challenge the scholars' assertion, rather than proving no earlier unnoticed cases exist. The existence of a single valid counterexample suffices.

Without assuming \textbf{A}, the argument falls apart because the playing cards might not constitute ``paper currency.'' Thus, the conclusion is properly drawn only if \textbf{A} is assumed.

\textbf{Answer:} (A) The playing cards in use in Quebec in the late 1600s were made of paper.
\end{arespbox}

\modelhdr{DeepSeek-R1 Distill 7B}{(DeepSeek, 7B, distilled from DeepSeek-R1)}
\begin{athinkboxweak}[ $\cdot$ 6{,}990 chars]
Okay, so I've got this LSAT Logical Reasoning question here, and I need to figure out the answer. Let me read it carefully.

The researcher is saying that some scholars think the first government other than China to issue paper currency was Massachusetts in 1690. But the researcher argues these scholars are wrong. The reasoning given is about New France in 1685. During a coin shortage, they couldn't use coins, so they used playing cards as currency. These cards were used like coins but could be converted to coins later.

The question is asking what assumption is needed for the researcher's conclusion to be properly drawn. The options are A to E.

Alright, let's break this down. The researcher is saying that since New France used playing cards as a substitute for coins (paper currency, in this case), Massachusetts can't be the first. So the researcher is using the example of New France to show that Massachusetts's use of paper currency wasn't the first.

Wait, but the scholars are saying that Massachusetts was the first, and the researcher is countering that. So the researcher is arguing that the use of paper currency in New France before Massachusetts shows that Massachusetts didn't have it first. Therefore, the scholars' claim is incorrect.

So, what's the researcher's main point? They're saying that the use of paper currency in New France was before Massachusetts, so the first government was actually New France, not Massachusetts.

Now, the question is about the assumption required for the researcher's conclusion to be valid. So, the researcher is making an argument based on the New France example. What assumption must hold true for this argument to be valid?

Looking at the options:

(A) The playing cards in use in Quebec in the late 1600s were made of paper.

Hmm, the example given is from 1685 in New France. So the late 1600s would be later than that. But the argument is about the first government to issue paper currency. If the playing cards in 1685 were paper, that would support the claim. But the question is about the late 1600s, which is after 1685. Maybe the assumption is that the cards were paper, but I'm not sure.

(B) The Chinese government did not issue paper currency during the 1600s.

Wait, the researcher is comparing New France's use of paper currency to Massachusetts, not to China. The Chinese government isn't directly part of the argument. So why would this be an assumption? The researcher doesn't mention China in his argument, so this might not be necessary.

(C) Playing cards were not available to inhabitants of the Massachusetts colony before 1690.

So, the researcher is saying that since in 1685, before 1690, New France used playing cards as currency, and Massachusetts didn't have it before 1690, then Massachusetts's use in 1690 wasn't the first. This option is saying that playing cards weren't available in Massachusetts before 1690. If that were true, then the researcher's point holds. But wait, the question is what's assumed for the conclusion to be properly drawn. So the researcher is assuming that playing cards weren't used in Massachusetts before 1690. If they were used, then the first use could have been earlier. So yes, this seems like a necessary assumption.

(D) The Massachusetts colony's issue of paper currency was not prompted by a coin shortage.

The researcher is saying that New France ran into a coin shortage and used playing cards. He's implying that Massachusetts's use of paper currency wasn't because of a shortage. But the researcher's main argument is about the first use, not necessarily why Massachusetts used paper currency. So I'm not sure if this is an assumption. Maybe it's more about the fact that the shortage in New France is used to show that Massachusetts's use isn't the first. But the assumption might not require this.

(E) Every government that issued paper currency in the 1600s left records pertaining to that currency.

This would be relevant if the researcher is relying on evidence from New France to prove Massachusetts wasn't first. If not every government left records, then the evidence might not be reliable. But the researcher is directly citing the New France example, so maybe he doesn't need this assumption. However, the question is about the conclusion being properly drawn, so if there's no record, the argument might not hold. But I'm not sure how necessary this is.

Wait, let me think again. The researcher is using the example of New France in 1685 to say that before 1690, someone else had used paper currency. So he's assuming that the playing cards in New France were a form of paper currency, and that this happened before 1690. So, for the argument to hold, we need to assume that playing cards were paper currency and that they were used before Massachusetts's 1690 issue.

So, option C is saying that playing cards weren't available in Massachusetts before 1690. That would mean that the first use was indeed in New France in 1685, making the Massachusetts 1690 not the first. So yes, that's an assumption.

Option A talks about late 1600s, but the example is from 1685. If the cards were paper in 1685, that supports the claim, but the late 1600s is after 1685. So unless the assumption is that the cards were paper in 1685, which is before 1690, but the option says late 1600s, which is later. Wait, no, the late 1600s would be after 1600, like 1660s or 1670s. So if the playing cards were paper in the late 1600s, that's before 1685, which is when New France used them. So that might not be necessary. The key is that the playing cards were used as paper currency in 1685, so they were paper.

So, I think option C is the right answer because the researcher needs to assume that playing cards weren't used in Massachusetts before 1690 to make the point that New France was earlier. If playing cards were used in Massachusetts before 1690, then that would contradict the argument.

Wait, but the researcher is saying that the first government other than China was New France in 1685 using playing cards. So, the researcher is assuming that playing cards in New France were a form of paper currency, and that this happened before 1690. So, he's assuming that the playing cards were indeed paper currency and not something else, and that they were used before 1690.

So, option C is that playing cards weren't available in Massachusetts before 1690. If that were true, then the first use was in New France. So yes, that's an assumption.

Option A is about late 1600s, but the example is from 1685, so it's not necessary. The playing cards in 1685 were paper, but the late 1600s is a different time frame. So, unless the playing cards were made of paper in the late 1600s, which is before 1685, that's not necessary. The researcher's argument is about 1685, so maybe the assumption is about the late 1600s, but I think that's not necessary. The key is about the availability in Massachusetts before 1690.

Therefore, the correct answer is C.
\end{athinkboxweak}

\begin{arespboxweak}[ $\cdot$ Answer: (C) \wrong]
The correct answer is (C). The researcher assumes that playing cards were not available in the Massachusetts colony before 1690. This assumption is necessary because if playing cards were used in Massachusetts before 1690, it would contradict the argument that the first government to issue paper currency was New France in 1685. Therefore, the assumption allows the researcher to conclude that Massachusetts's 1690 issue of paper currency was not the first.
\end{arespboxweak}

\modelhdr{DeepSeek-R1 Distill Llama 8B}{(DeepSeek, 8B, distilled from DeepSeek-R1)}
\begin{athinkboxweak}[ $\cdot$ 26{,}514 chars]
Okay, so I've got this LSAT Logical Reasoning question here, and I need to figure out the answer. Let me take it step by step. 

First, let's read the passage carefully. The Researcher is arguing against some scholars who say that Massachusetts was the first government to issue paper currency in 1690. The Researcher counters by pointing out that in 1685, there was a coin shortage in New France, and they resorted to using playing cards as a form of payment. These cards could be exchanged for coins once they arrived from France. 

So, the Researcher is saying that the Massachusetts colony's claim isn't correct because another government (New France) had already issued a form of paper currency earlier---specifically, through playing cards in 1685. Therefore, the conclusion is that the scholars are incorrect because they didn't account for this earlier example.

Now, the question is about what is assumed in order for the Researcher's conclusion to be properly drawn. The options are from (A) to (E).

Let me break down the logical structure here. The Researcher's argument is of the form:

1. Some scholars say X (Massachusetts was first to issue paper currency in 1690).
2. But in reality, Y (New France used playing cards as currency in 1685) happened.
3. Therefore, the scholars are incorrect.

This is a conditional or indirect argument. The conclusion is that the scholars are wrong, so we need to see what assumption is made to support that conclusion.

In logical terms, the Researcher is asserting that Y is a true premise that makes the conclusion valid. So, the assumption must be that the scholars' claim (X) is false because Y actually happened earlier.

Looking at the options:

(A) The playing cards in use in Quebec in the late 1600s were made of paper.

Wait, playing cards being made of paper isn't necessarily implied by the passage. The passage just says that the cards were used as pay and could be converted to coins. It doesn't specify the material. So, this might not be the assumption.

(B) The Chinese government did not issue paper currency during the 1600s.

This is interesting. The passage doesn't mention China at all. It only talks about New France and Massachusetts. So, bringing in China seems like a distraction or perhaps a red herring. Unless the assumption is about excluding other possibilities, but I don't think this is the right path.

(C) Playing cards were not available to inhabitants of the Massachusetts colony before 1690.

The passage doesn't mention anything about Massachusetts using playing cards. It's about New France using them. So, unless this is a counter-factual, I don't see how this is an assumption. It seems more like a statement about Massachusetts, which isn't in the passage.

(D) The Massachusetts colony's issue of paper currency was not prompted by a coin shortage.

This seems relevant. The Researcher's argument is that New France used playing cards because of a coin shortage, implying that Massachusetts's paper currency was also due to a coin shortage. But if we assume that the Massachusetts colony wasn't prompted by a coin shortage, then the argument would be that their paper currency wasn't necessary because there was no shortage. But wait, the Researcher is arguing against the scholars' claim that Massachusetts was first. So, the assumption here would be that the scholars didn't consider New France's earlier example, which was due to a coin shortage. Therefore, to make the conclusion that the scholars are incorrect, we need to assume that the Massachusetts paper currency was also due to a similar situation. But option D says the opposite---that it wasn't prompted by a coin shortage. Hmm, maybe I need to think again.

Wait, no. The Researcher is pointing out that New France used playing cards because of a coin shortage, which contradicts the scholars' claim that Massachusetts was first. So, the assumption is that the scholars didn't know about New France's earlier use. But the question is about what is assumed for the conclusion to be properly drawn, i.e., the scholars' conclusion is incorrect because of the New France example. So, maybe the assumption is that the Massachusetts colony did have a coin shortage, which would make the scholars' claim incorrect. But none of the options directly state that. Option D says the opposite, so perhaps D is incorrect.

(E) Every government that issued paper currency in the 1600s left records pertaining to that currency.

This is a statement about the behavior of governments issuing paper currency. The passage mentions that New France used playing cards as currency, but it doesn't say whether they kept records about it. So, the assumption here would be that the scholars' claim (that Massachusetts was first) is based on records, but since New France didn't leave records (or perhaps didn't have the same), this assumption is being challenged. But the passage doesn't specify whether Massachusetts left records or not, so it's unclear.

Wait, let me think again. The Researcher's conclusion is that the scholars are incorrect because New France had a paper currency earlier. So, the assumption must be that the scholars' claim that Massachusetts was the first is false because there was an earlier instance. Therefore, the assumption is that the scholars didn't consider this earlier instance. But looking at the options, none directly state that.

Alternatively, perhaps the assumption is that the scholars assumed that the first paper currency was in 1690, but the Researcher is showing that it was earlier. So, the key is that the scholars made a mistake in their assumption that Massachusetts was the first. Therefore, the assumption they made was that no earlier government issued paper currency, which the Researcher is disproving by pointing to New France.

But the question is about what is assumed for the Researcher's conclusion to be properly drawn. So, the Researcher's conclusion is that the scholars are incorrect because New France did it earlier. Therefore, the assumption is that the scholars didn't know about New France's paper currency.

Looking at the options, (E) says that every government that issued paper currency in the 1600s left records. If the assumption is that the scholars thought Massachusetts was the first because they had records, but New France didn't have records, then (E) would be the assumption. Wait, but the passage doesn't say anything about records. It just mentions that New France used playing cards. So, maybe (E) is the assumption because the Researcher is implying that the scholars didn't consider New France because they didn't have records, whereas Massachusetts did.

Alternatively, perhaps (D) is the assumption. Let me reconsider. If the assumption is that the Massachusetts colony's paper currency was because of a coin shortage, then the scholars' claim that they were the first would be incorrect because New France had already done it in 1685 due to a coin shortage. So, the assumption is that the coin shortage in Massachusetts led them to issue paper currency, similar to New France. But the passage doesn't explicitly say that Massachusetts had a coin shortage. It just mentions New France. So, if we assume that Massachusetts also had a coin shortage, then the scholars' claim is wrong because they didn't consider New France's earlier example.

But the options don't mention Massachusetts having a coin shortage. Option (D) says that the Massachusetts colony's issue wasn't prompted by a coin shortage. If that's the assumption, then the scholars would have thought it was for another reason, but the passage doesn't state that. So, perhaps (D) is the opposite of what is needed.

Alternatively, if the assumption is that New France's use of playing cards was because of a coin shortage, then the scholars' claim about Massachusetts would be incorrect because they didn't consider this earlier instance. But again, the options don't directly state that.

Wait, perhaps the key is that the scholars assumed that the first paper currency was in 1690, and the Researcher is showing that it was earlier. So, the assumption is that the scholars didn't know about earlier instances. Therefore, the assumption is that the scholars didn't consider the possibility of other governments issuing paper currency before 1690. But none of the options state that. However, looking at the options again:

(A) is about the material of the cards, which the passage doesn't specify.

(B) is about China, which is irrelevant.

(C) is about Massachusetts not having playing cards before 1690, which isn't relevant.

(D) is about whether Massachusetts's paper currency was due to a coin shortage, which might be an assumption.

(E) is about whether every government that issued paper currency left records.

Wait, the Researcher's argument is that New France issued paper currency earlier, so the assumption must be that the scholars didn't consider this. Therefore, the assumption is that the scholars thought Massachusetts was the first. So, the assumption is that the scholars made a mistake by not considering New France, which is an example of prior use. Therefore, in terms of logical structure, the conclusion is that the scholars are wrong because there's prior evidence. So, the assumption is that the scholars didn't know about prior cases, meaning their claim was based on an incomplete understanding.

But looking at the options, (E) is about records. If the assumption is that the scholars thought Massachusetts was the first because they had records, but New France didn't have records, then (E) is the assumption. But the passage doesn't mention records, so perhaps (E) is the correct answer.

Alternatively, (D) is about whether Massachusetts's paper currency was due to a coin shortage. If the assumption is that the scholars thought it wasn't due to a coin shortage, but the Researcher is showing that it was, then (D) is the opposite. So, perhaps (D) is the assumption that the scholars didn't realize.

Wait, I'm getting confused. Let me think about the logical structure again.

The Researcher's argument is:

- Scholars claim A: Massachusetts was first (1690).
- But in reality, B: New France used paper currency in 1685.
- Therefore, A is incorrect.

So, the conclusion is not A, but the assumption is that B is true, which contradicts A.

In logical terms, the Researcher is saying that A is false because B is true. So, the assumption is that B is true, and A is false because of B.

Therefore, the assumption is that the scholars didn't consider B when they made claim A. So, the assumption is that the scholars assumed that no prior government issued paper currency.

But looking at the options, none directly state that. Therefore, the next best thing is to see which option is implied by the conclusion.

The passage mentions that New France used playing cards as currency because of a coin shortage. So, the key fact is that there was a coin shortage leading to the use of paper currency. Therefore, the assumption is that the scholars didn't consider this fact---i.e., they didn't know about New France's paper currency because of a coin shortage, which would have preceded Massachusetts's 1690 issue.

So, the assumption is that the scholars thought Massachusetts was the first because they didn't know about New France's earlier instance, which was due to a coin shortage.

Therefore, the assumption is that the scholars didn't consider the possibility of other governments issuing paper currency before 1690, specifically New France due to a coin shortage.

Looking at the options, (D) says that the Massachusetts colony's issue wasn't prompted by a coin shortage. If that's the case, then the scholars might have thought it was for another reason, but the passage shows that New France was because of a coin shortage. So, if we assume that Massachusetts was also due to a coin shortage, then the scholars' claim is incorrect. But (D) says the opposite. So, perhaps (D) is the assumption that the scholars didn't realize it was due to a coin shortage, making their claim incorrect.

Alternatively, if the assumption is that the scholars didn't know about New France, which was due to a coin shortage, then (D) is not necessarily the assumption. This is getting a bit tangled.

Wait, another approach: The conclusion is that the scholars are incorrect. So, the assumption must be that their reasoning is flawed, i.e., they didn't consider all possibilities. Therefore, the assumption is that the scholars didn't know about New France's paper currency. So, the assumption is that they didn't have information about other governments issuing paper currency before 1690.

Looking at the options, (E) is about whether every government that issued paper currency left records. If the assumption is that the scholars thought Massachusetts was the first because they had records, but New France didn't have records, then (E) is the assumption. But the passage doesn't mention records, so perhaps (E) is the answer.

Alternatively, (D) is about whether Massachusetts's paper currency was due to a coin shortage. If the assumption is that it wasn't, then the scholars would have a reason to think Massachusetts was first, but the passage shows that New France was due to a coin shortage, implying that Massachusetts might have been too. Therefore, assuming (D) is incorrect, the scholars' claim is wrong.

Wait, I'm overcomplicating this. Let me think about what is needed for the conclusion to properly draw. The conclusion that the scholars are incorrect is based on the fact that New France issued paper currency earlier. Therefore, the assumption is that the scholars didn't consider this fact. So, the assumption is that the scholars thought Massachusetts was the first. Therefore, the assumption is that the scholars' claim is based on the idea that no prior government issued paper currency.

But the options don't include that directly. So, perhaps the closest is (E), which is about records. If the assumption is that the scholars thought Massachusetts was the first because they had records, but New France didn't have records, then (E) would be the assumption. However, the passage doesn't mention records, so this might be a stretch.

Alternatively, perhaps (D) is the answer. If the assumption is that the Massachusetts colony's paper currency was not due to a coin shortage, then the scholars might have thought that Massachusetts was the first because they didn't realize it was for the same reason as New France. But the passage shows that New France was due to a coin shortage, so if Massachusetts's paper currency was also due to a coin shortage, then the scholars were incorrect in thinking it was first. Therefore, the assumption is that the scholars didn't realize that both New France and Massachusetts were issuing paper currency due to coin shortages, hence the need to assume that Massachusetts was also due to a coin shortage.

But again, the options don't state that. Option (D) says it wasn't prompted by a coin shortage, which would make the assumption that the scholars thought it was for another reason, but the passage shows it was due to a coin shortage, so the scholars were wrong.

Wait, perhaps the key is that the scholars made an assumption that Massachusetts was the first, but the Researcher is showing that New France was first. So, the assumption is that the scholars assumed that no prior government issued paper currency. Therefore, the assumption is that they didn't know about New France.

Looking back at the options, (E) is about records. If the assumption is that the scholars thought Massachusetts was the first because they had records, but New France didn't, then (E) would be the assumption. Since the passage doesn't mention records, but the argument is that the scholars didn't consider New France's example, it might be that the scholars assumed that Massachusetts was the first because they had the records, while other governments didn't keep records, hence (E) is the assumption.

Alternatively, perhaps (D) is the assumption. If the scholars thought that Massachusetts was the first because they didn't have a coin shortage, then the assumption is that the coin shortage wasn't the reason. But the passage shows that New France was due to a coin shortage, so if Massachusetts's paper currency was also due to a coin shortage, then the scholars were wrong. Therefore, the assumption is that the scholars didn't realize it was due to a coin shortage, making (D) the assumption that it wasn't.

But I'm still not entirely sure. Let me try to approach it differently. The question is asking what is assumed for the conclusion to be properly drawn. So, the conclusion is that the scholars are incorrect because New France issued paper currency earlier. Therefore, the assumption is that the scholars didn't know about New France's paper currency.

Looking at the options, (E) is about whether every government that issued paper currency left records. If the scholars assumed that the first paper currency was issued by Massachusetts because they thought only Massachusetts had records, then (E) is the assumption. But since the passage doesn't mention records, it's a bit of a stretch.

Alternatively, perhaps (D) is the assumption. If the scholars thought that Massachusetts was the first because they didn't have a coin shortage, and the Researcher is showing that New France did have a coin shortage and used paper currency, then the assumption is that the scholars didn't realize that both were due to coin shortages. Therefore, the assumption is that the scholars thought Massachusetts wasn't due to a coin shortage.

Wait, but the passage says that New France used playing cards because of a coin shortage. It doesn't say anything about Massachusetts. So, the assumption must be that the scholars didn't consider New France, which was due to a coin shortage. Therefore, the assumption is that the scholars didn't know about New France's paper currency, making their claim incorrect.

But looking at the options, none directly state that. So, perhaps the best answer is (E), which is about records. Because if the assumption is that the scholars thought Massachusetts was the first because they had records, while New France didn't, then (E) would be the assumption.

Alternatively, if the assumption is that the scholars thought Massachusetts was the first because they didn't realize other governments had issued paper currency, then (E) would be the assumption.

Wait, I think I'm overcomplicating it. Let me look at each option again:

(A) Playing cards in Quebec were made of paper. The passage doesn't specify, so this is an assumption beyond the passage.

(B) Chinese government didn't issue paper currency in 1600s. The passage doesn't mention China, so this is irrelevant.

(C) Playing cards weren't available in Massachusetts before 1690. The passage doesn't talk about Massachusetts using playing cards, so this is irrelevant.

(D) Massachusetts's paper currency wasn't prompted by a coin shortage. If this is the assumption, then the scholars thought it was for another reason, but the passage shows New France was due to a coin shortage, so if Massachusetts was also due to a coin shortage, the scholars were wrong. Therefore, assuming (D) is incorrect, the conclusion follows.

(E) Every government that issued paper currency left records. If the assumption is that the scholars thought Massachusetts was the first because they had records, but New France didn't, then (E) is the assumption.

But the passage doesn't mention records, so perhaps (E) is the answer because it's about whether they left records, which ties into the scholars' assumption.

Wait, the Researcher's conclusion is that the scholars are incorrect because New France had paper currency earlier. Therefore, the assumption is that the scholars didn't consider this fact, meaning they didn't know about it. So, the assumption is that the scholars made a mistake by not considering New France's example. Therefore, the assumption is that the scholars thought Massachusetts was the first.

But looking at the options again, perhaps the best answer is (E) because it relates to the idea that the scholars might have thought Massachusetts was the first because they had records, while New France didn't. So, the assumption is that the scholars assumed that only Massachusetts had paper currency records, hence they thought it was first.

Alternatively, (D) is about whether the paper currency in Massachusetts was due to a coin shortage. If the assumption is that it wasn't, then the scholars would have thought it was for another reason, making their claim incorrect. But the passage shows that New France was due to a coin shortage, so if Massachusetts was also due to a coin shortage, the scholars were wrong. Therefore, the assumption is that the scholars didn't realize it was due to a coin shortage, making (D) the assumption.

But I'm still not 100\% sure. Let me try to think about how the conclusion is drawn. The conclusion is that the scholars are incorrect because New France had paper currency earlier. Therefore, the assumption is that the scholars didn't consider this fact, so they thought Massachusetts was first. Therefore, the assumption is that the scholars didn't know about New France, which is an earlier example.

Looking at the options, (E) is about records. If the assumption is that the scholars thought Massachusetts was the first because they had records, but New France didn't, then (E) is the assumption. Since the passage doesn't mention records, but the argument is about prior examples, (E) seems to fit because it's about the availability of records, which affects whether the scholars knew about earlier instances.

Therefore, I think the answer is (E). But I'm still a bit uncertain. Another angle: The Researcher is using a counterexample to disprove the scholars' claim. Therefore, the assumption is that the scholars didn't know about this counterexample, which is New France in 1685. So, the assumption is that the scholars didn't consider the possibility of other governments issuing paper currency before 1690. Therefore, the assumption is that the scholars thought Massachusetts was the first without considering other possibilities.

Looking at the options, (E) is about records. If the assumption is that the scholars thought Massachusetts was the first because they had records, while New France didn't, then (E) is the assumption. Therefore, the answer is (E).

Wait, no. Let me clarify. The passage doesn't mention records. So, perhaps (E) is the assumption because the Researcher is implying that the scholars didn't consider New France's paper currency, which might have been because New France didn't leave records, while Massachusetts did. So, the assumption is that the scholars assumed Massachusetts was the first based on their records, while New France didn't have records. Therefore, the answer is (E).

Alternatively, perhaps (D) is the answer because the scholars assumed that Massachusetts's paper currency wasn't due to a coin shortage, hence they thought it was first. But since New France was due to a coin shortage, the scholars were wrong. Therefore, the assumption is that the scholars didn't realize that paper currency could be issued due to a coin shortage, making (D) the assumption.

I think I'm going in circles here. Let me try to sum up.

The Researcher's conclusion is that the scholars are incorrect because New France issued paper currency earlier. Therefore, the assumption is that the scholars didn't consider this fact, meaning they thought Massachusetts was first without knowing about New France.

Looking at the options, (E) is about records, which ties into the idea that the scholars might have thought Massachusetts was the first because they had records, while New France didn't. Therefore, the assumption is (E).

Alternatively, (D) is about whether the paper currency in Massachusetts was due to a coin shortage. If the assumption is that it wasn't, then the scholars would have thought it was first for another reason. But since New France was due to a coin shortage, the scholars were incorrect, so the assumption is that the scholars didn't realize it was due to a coin shortage, making (D) the assumption.

But I think (E) is more about the scholars' knowledge of records, which would affect their conclusion. So, I'm leaning towards (E).

Wait, no. The passage doesn't mention records, so perhaps (E) isn't the right answer. Therefore, maybe (D) is the correct assumption. Because the passage mentions that New France was unable to use coins due to a shortage and had to use playing cards. Therefore, the assumption is that the scholars thought that paper currency started in 1690 in Massachusetts, not realizing that it had already happened in New France due to a coin shortage. Therefore, the assumption is that the paper currency in Massachusetts was not due to a coin shortage, but the passage shows it was, so the assumption is that the scholars didn't realize it was due to a coin shortage.

Therefore, the answer is (D). Because the Researcher is showing that New France used paper currency due to a coin shortage, implying that the scholars didn't consider this, so the assumption is that the scholars thought Massachusetts's paper currency wasn't due to a coin shortage, making their claim incorrect.

Wait, but the passage doesn't explicitly say that Massachusetts had a coin shortage. It only says that New France did. So, the assumption is that the scholars didn't realize that paper currency could be issued due to a coin shortage, hence they thought Massachusetts was first. Therefore, the assumption is that the scholars didn't consider the possibility of other governments issuing paper currency due to a coin shortage, which is a broader assumption.

But among the options, (D) is the only one related to a coin shortage. Therefore, I think the answer is (D). Because the assumption is that the scholars didn't realize that paper currency could be issued due to a coin shortage, making their claim that Massachusetts was the first incorrect.

Alternatively, if the assumption is that the scholars didn't know about New France's paper currency, then (B) or (E) could be the answer. But since the passage doesn't mention China or records, (B) and (E) are less likely. Therefore, (D) is more plausible.

So, after all this, I think the answer is (D).
\end{athinkboxweak}

\begin{arespboxweak}[ $\cdot$ Answer: (D) \wrong]
The correct answer is (D). The Researcher's argument assumes that the Massachusetts colony's issue of paper currency was prompted by a coin shortage, which contradicts the scholars' claim that they were the first. However, upon closer examination, the assumption is that the scholars didn't realize this connection, meaning they thought it wasn't due to a coin shortage. Therefore, the answer is:

(D) The Massachusetts colony's issue of paper currency was not prompted by a coin shortage.

But wait, no. The passage doesn't mention Massachusetts having a coin shortage. It only mentions New France. Therefore, the assumption must be that the scholars didn't consider New France's earlier example, which was due to a coin shortage. Therefore, the assumption is that the scholars didn't realize that paper currency could be issued due to a coin shortage, making their claim incorrect. So, the answer is (D).

Wait, but the passage doesn't say that Massachusetts had a coin shortage. It only says New France did. Therefore, the assumption is that the scholars didn't consider New France's example, which was due to a coin shortage. Therefore, the assumption is that the scholars didn't realize that paper currency could be issued due to a coin shortage, making their claim incorrect. So, the answer is (D).

But the passage doesn't explicitly state that Massachusetts had a coin shortage, so perhaps the answer is (E). The assumption is that the scholars thought Massachusetts was the first because they had records, while New France didn't. So, the answer is (E).

I'm still unsure. Let me conclude.

The Researcher's conclusion is based on the fact that New France issued paper currency earlier due to a coin shortage. Therefore, the assumption is that the scholars didn't consider this, meaning they thought Massachusetts was first. The closest option is (D), assuming that the scholars didn't realize it was due to a coin shortage.

But since the passage doesn't mention Massachusetts having a coin shortage, perhaps (E) is the answer, assuming that the scholars thought Massachusetts had records while New France didn't.

I think I've spent too much time on this, but I'll go with (D) as the answer because it relates to the coin shortage, which is key to the argument.

\textbf{Final Answer:} (D)
\end{arespboxweak}

\section{Prompt Templates}
\label{app:prompts}

\newtcolorbox{promptbox}[1][]{
  enhanced,
  colback=gray!8, colframe=gray!55, boxrule=0.5pt, arc=2pt,
  left=6pt, right=6pt, top=8pt, bottom=5pt,
  title={\scriptsize\bfseries #1},
  colbacktitle=gray!18, coltitle=black,
  fonttitle=\scriptsize\bfseries,
  attach boxed title to top left={yshift=-2pt, xshift=8pt},
  boxed title style={colback=gray!18, colframe=gray!55, boxrule=0.4pt, arc=1pt,
                     left=4pt, right=4pt, top=1.5pt, bottom=1.5pt}
}

This appendix presents the full text of the prompt templates used in the prompt-sensitivity experiments of Section~\ref{sec:meth_prompt}, with system prompts and user messages reproduced verbatim for each of the three conditions across both Logical Reasoning and Reading Comprehension.

\subsection{Condition A (Minimal Prompting)}

No system prompt. The model receives only the user message.

\begin{promptbox}[Condition A --- User Message (LR)]
\small\ttfamily
Here is an LSAT Logical Reasoning question. What is the answer?\par
\vspace{4pt}
\{stimulus\}\par
\vspace{4pt}
\{question\}\par
\vspace{4pt}
(A) \{A\}\\ (B) \{B\}\\ (C) \{C\}\\ (D) \{D\}\\ (E) \{E\}
\end{promptbox}

\begin{promptbox}[Condition A --- User Message (RC)]
\small\ttfamily
Here is an LSAT Reading Comprehension passage and question. What is the answer?\par
\vspace{4pt}
\{stimulus\}\par
\vspace{4pt}
\{question\}\par
\vspace{4pt}
(A) \{A\}\\ (B) \{B\}\\ (C) \{C\}\\ (D) \{D\}\\ (E) \{E\}
\end{promptbox}

\subsection{Condition B (Structured Prompting)}

\begin{promptbox}[Condition B --- System Prompt (LR)]
\small\ttfamily
You are a world-class LSAT instructor who has scored a perfect 180. You approach every Logical Reasoning question with the following method:\par
1. Identify the conclusion and premises of the argument.\par
2. Determine the reasoning pattern (e.g., causal, conditional, analogy).\par
3. Identify the gap, assumption, or flaw in the reasoning.\par
4. Eliminate each wrong answer choice with a specific reason.\par
5. Confirm your selected answer by verifying it directly addresses the question stem.\par
\vspace{4pt}
Think step by step. Be thorough and precise. After your analysis, clearly state your final answer as "Answer: (X)".
\end{promptbox}

\begin{promptbox}[Condition B --- System Prompt (RC)]
\small\ttfamily
You are a world-class LSAT instructor who has scored a perfect 180. You approach every Reading Comprehension question with the following method:\par
1. Identify the main thesis, structure, and purpose of the passage.\par
2. Note the author's tone and attitude toward the subject matter.\par
3. Locate the specific lines or paragraphs relevant to the question.\par
4. Eliminate each wrong answer choice with a specific reason.\par
5. Confirm your selected answer by verifying it is directly supported by the passage text.\par
\vspace{4pt}
Think step by step. Be thorough and precise. After your analysis, clearly state your final answer as "Answer: (X)".
\end{promptbox}

\noindent User message is identical to Condition A for both LR and RC.

\subsection{Condition C (Constrained Prompting)}

\begin{promptbox}[Condition C --- System Prompt (LR and RC)]
\small\ttfamily
Respond with ONLY the letter of the correct answer (A, B, C, D, or E). Nothing else.
\end{promptbox}

\noindent User message is identical to Condition A for both LR and RC.

\section{Process Reward Model Training Configuration}
\label{app:prm}

This appendix details the configuration of the self-PRM developed in the process-supervision pilot of Section~\ref{sec:meth_prm}, covering base model selection, QLoRA hyperparameters, training settings, the inference-time scoring prompt, and parse-failure handling.

\subsection{Base Model and Adapter}

The self-PRM is built on \texttt{DeepSeek-R1-Distill-Qwen-7B} (7.6B parameters) using QLoRA \citep{dettmers2023qlora} with the following configuration.

\begin{table}[h]
\centering
\small
\begin{tabular}{ll}
\toprule
Parameter & Value \\
\midrule
Quantization & 4-bit (NF4) \\
LoRA rank ($r$) & 8 \\
LoRA alpha & 16 \\
LoRA dropout & 0 \\
Target modules & q\_proj, k\_proj, v\_proj, o\_proj, gate\_proj, up\_proj, down\_proj \\
Trainable parameters & 20,185,088 / 7,635,801,600 (0.26\%) \\
\bottomrule
\end{tabular}
\caption{QLoRA adapter configuration.}
\label{tab:qlora_config}
\end{table}

\subsection{Training}

Fine-tuning was performed using the Unsloth library with the \texttt{SFTTrainer} from TRL on a single NVIDIA A100 GPU.

\begin{table}[h]
\centering
\small
\begin{tabular}{ll}
\toprule
Parameter & Value \\
\midrule
Training examples & 209 (PT 140--141) \\
Epochs & 5 \\
Learning rate & $2 \times 10^{-4}$ \\
Scheduler & Cosine \\
Warmup steps & 10 \\
Batch size per device & 1 \\
Gradient accumulation steps & 8 \\
Effective batch size & 8 \\
Max sequence length & 2048 tokens \\
Precision & FP16 \\
Optimizer & AdamW (8-bit) \\
Total training steps & 135 \\
Training time & $\sim$71 minutes \\
Final training loss & 0.361 \\
\bottomrule
\end{tabular}
\caption{Training hyperparameters.}
\label{tab:training_config}
\end{table}

Training loss decreased monotonically from 1.52 (step 5) to 0.36 (step 135), indicating stable convergence with no signs of divergence or overfitting on the small dataset.

\subsection{Scoring Prompt}

At inference time, the self-PRM receives a system prompt instructing it to evaluate the reasoning rigor of a thinking trace. The system prompt is:

\begin{promptbox}[Self-PRM Scoring Prompt]
\small\ttfamily
You are an expert LSAT reasoning evaluator. You will be given an LSAT question and a model's reasoning process (thinking trace).\par
\vspace{4pt}
Evaluate the quality of the reasoning and predict how likely it is to produce the correct answer.\par
\vspace{4pt}
Assess: (1) Does the reasoning correctly identify the argument structure? (2) Does it correctly evaluate each answer choice? (3) Is the final answer justified by sound reasoning?\par
\vspace{4pt}
Respond in JSON format only: \{``score'': <float 0.0--1.0>, ``best\_answer'': ``<A--E>'', ``justification'': ``<1--2 sentences>''\}
\end{promptbox}

The user message contains the full LSAT question followed by the model's thinking trace. Output is parsed as JSON to extract the predicted score.

\clearpage

\section{Prompt Sensitivity: Per-Section Detailed Results}
\label{app:prompt_section}

This appendix expands the prompt-sensitivity analysis of Section~\ref{sec:meth_prompt}, supplementing Tables~\ref{tab:prompt_160}--\ref{tab:prompt} with separate Logical Reasoning and Reading Comprehension accuracies under each of the three prompt conditions, reported on both the Official Test (Table~\ref{tab:prompt_section_160}) and PT 150--159 (Table~\ref{tab:prompt_section}).

\vfill

\begin{table}[H]
\caption{Prompt sensitivity by section on the Official Test ($N=77$; LR $N=50$, RC $N=27$). Cochran's Q across three conditions.}
\label{tab:prompt_section_160}
\centering
\small
\begin{tabular}{llccccc}
\toprule
Model & Section & Cond A & Cond B & Cond C & $Q(2)$ & $p$ \\
\midrule
\multirow{3}{*}{GPT-5} & Overall & 98.7 & 98.7 & 98.7 & 0.00 & 1.000 \\
& LR & 98.0 & 98.0 & 98.0 & 0.00 & 1.000 \\
& RC & 100.0 & 100.0 & 100.0 & 0.00 & 1.000 \\
\cmidrule(lr){1-7}
\multirow{3}{*}{Claude Opus 4} & Overall & 98.7 & 96.1 & 96.1 & 2.67 & .264 \\
& LR & 98.0 & 94.0 & 94.0 & 2.67 & .264 \\
& RC & 100.0 & 100.0 & 100.0 & 0.00 & 1.000 \\
\cmidrule(lr){1-7}
\multirow{3}{*}{Gemini 2.5 Pro} & Overall & 96.1 & 96.1 & 98.7 & 2.67 & .264 \\
& LR & 94.0 & 94.0 & 98.0 & 2.67 & .264 \\
& RC & 100.0 & 100.0 & 100.0 & 0.00 & 1.000 \\
\cmidrule(lr){1-7}
\multirow{3}{*}{DeepSeek-R1} & Overall & 98.7 & 100.0 & 100.0 & 2.00 & .368 \\
& LR & 100.0 & 100.0 & 100.0 & 0.00 & 1.000 \\
& RC & 96.3 & 100.0 & 100.0 & 2.00 & .368 \\
\cmidrule(lr){1-7}
\multirow{3}{*}{Kimi K2 Thinking} & Overall & 100.0 & 98.7 & 96.1 & 3.50 & .174 \\
& LR & 100.0 & 98.0 & 96.0 & 2.00 & .368 \\
& RC & 100.0 & 100.0 & 96.3 & 2.00 & .368 \\
\cmidrule(lr){1-7}
\multirow{3}{*}{QwQ-32B} & Overall & 92.2 & 93.5 & 93.5 & 0.40 & .819 \\
& LR & 92.0 & 92.0 & 92.0 & 0.00 & 1.000 \\
& RC & 92.6 & 96.3 & 96.3 & 0.67 & .717 \\
\midrule
\multirow{3}{*}{DS-R1 Distill 7B} & Overall & 55.8 & 50.6 & 50.6 & 1.00 & .607 \\
& LR & 54.0 & 44.0 & 44.0 & 1.92 & .382 \\
& RC & 59.3 & 63.0 & 63.0 & 0.33 & .846 \\
\cmidrule(lr){1-7}
\multirow{3}{*}{DS-R1 Distill Llama 8B} & Overall & 53.2 & 59.7 & 70.1 & 9.21 & .010 \\
& LR & 46.0 & 46.0 & 66.0 & 9.09 & .011 \\
& RC & 66.7 & 85.2 & 77.8 & 6.33 & .042 \\
\bottomrule
\end{tabular}
\end{table}
\vspace*{\stretch{1.3}}
\clearpage

\null\vfill
\begin{table}[H]
\caption{Prompt sensitivity by section on PT 150--159 ($N=1{,}037$; LR $N=633$, RC $N=404$). Cochran's Q across three conditions.}
\label{tab:prompt_section}
\centering
\small
\begin{tabular}{llccccccl}
\toprule
Model & Section & Cond A & Cond B & Cond C & $Q(2)$ & $p$ & TOST \\
\midrule
\multirow{3}{*}{GPT-5} & Overall & 98.8 & 98.6 & 98.6 & 0.93 & .627 & 3/3 EQUIV \\
& LR & 99.1 & 98.7 & 98.6 & 2.00 & .368 & 3/3 EQUIV \\
& RC & 98.5 & 98.5 & 98.5 & 0.00 & 1.000 & 3/3 EQUIV \\
\cmidrule(lr){1-8}
\multirow{3}{*}{Claude Opus 4} & Overall & 97.9 & 98.0 & 97.8 & 0.22 & .895 & 3/3 EQUIV \\
& LR & 97.9 & 97.8 & 97.6 & 0.40 & .819 & 3/3 EQUIV \\
& RC & 97.8 & 98.3 & 98.0 & 0.50 & .779 & 2/3 \\
\cmidrule(lr){1-8}
\multirow{3}{*}{Gemini 2.5 Pro} & Overall & 98.6 & 97.8 & 98.2 & 3.56 & .169 & 3/3 EQUIV \\
& LR & 99.1 & 97.5 & 98.1 & 9.50 & .009 & 1/3 \\
& RC & 97.8 & 98.3 & 98.3 & 0.73 & .695 & 1/3 \\
\cmidrule(lr){1-8}
\multirow{3}{*}{DeepSeek-R1} & Overall & 97.7 & 97.7 & 97.9 & 0.36 & .834 & 3/3 EQUIV \\
& LR & 97.8 & 98.3 & 98.1 & 1.27 & .529 & 3/3 EQUIV \\
& RC & 97.5 & 96.8 & 97.5 & 1.64 & .441 & 1/3 \\
\cmidrule(lr){1-8}
\multirow{3}{*}{Kimi K2 Thinking} & Overall & 97.2 & 96.5 & 97.0 & 1.28 & .528 & 3/3 EQUIV \\
& LR & 97.8 & 95.9 & 97.3 & 6.16 & .046 & 1/3 \\
& RC & 96.3 & 97.5 & 96.5 & 1.83 & .401 & 0/3 \\
\cmidrule(lr){1-8}
\multirow{3}{*}{QwQ-32B} & Overall & 93.2 & 93.3 & 93.2 & 0.07 & .967 & 2/3 \\
& LR & 92.9 & 92.4 & 91.9 & 0.98 & .612 & 0/3 \\
& RC & 93.8 & 94.8 & 95.0 & 1.24 & .539 & 0/3 \\
\midrule
\multirow{3}{*}{DS-R1 Distill 7B} & Overall & 56.7 & 57.9 & 56.6 & 0.79 & .674 & 0/3 \\
& LR & 54.0 & 54.0 & 53.9 & 0.01 & .996 & 0/3 \\
& RC & 60.9 & 63.9 & 60.9 & 1.89 & .388 & 0/3 \\
\cmidrule(lr){1-8}
\multirow{3}{*}{DS-R1 Distill Llama 8B} & Overall & 63.2 & 61.7 & 62.7 & 0.96 & .617 & 0/3 \\
& LR & 57.0 & 57.5 & 57.0 & 0.07 & .965 & 0/3 \\
& RC & 72.8 & 68.3 & 71.5 & 4.75 & .093 & 0/3 \\
\bottomrule
\end{tabular}
\end{table}
\vfill
\clearpage

\section{Position Bias: Per-Section Detailed Results}
\label{app:shuffle_section}

This appendix expands the position-bias analysis of Section~\ref{sec:meth_position}, supplementing Tables~\ref{tab:shuffle_160}--\ref{tab:shuffle} with separate Logical Reasoning and Reading Comprehension accuracies on original versus shuffled answer orderings, reported on both the Official Test (Table~\ref{tab:shuffle_section_160}) and PT 150--159 (Table~\ref{tab:shuffle_section}).

\vfill

\begin{table}[H]
\caption{Position bias by section on the Official Test ($N=77$; LR $N=50$, RC $N=27$). Original vs.\ shuffled (McNemar's exact test).}
\label{tab:shuffle_section_160}
\centering
\small
\begin{tabular}{llcccccc}
\toprule
Model & Section & Original & Shuffled & $b$ & $c$ & $p$ & $h$ \\
\midrule
\multirow{3}{*}{GPT-5} & Overall & 98.7 & 97.4 & 1 & 0 & 1.000 & $-$0.095 \\
& LR & 98.0 & 96.0 & 1 & 0 & 1.000 & $-$0.119 \\
& RC & 100.0 & 100.0 & 0 & 0 & 1.000 & 0.000 \\
\cmidrule(lr){1-8}
\multirow{3}{*}{Claude Opus 4} & Overall & 98.7 & 98.7 & 1 & 1 & 1.000 & 0.000 \\
& LR & 98.0 & 98.0 & 1 & 1 & 1.000 & 0.000 \\
& RC & 100.0 & 100.0 & 0 & 0 & 1.000 & 0.000 \\
\cmidrule(lr){1-8}
\multirow{3}{*}{Gemini 2.5 Pro} & Overall & 96.1 & 97.4 & 0 & 1 & 1.000 & +0.074 \\
& LR & 94.0 & 96.0 & 0 & 1 & 1.000 & +0.092 \\
& RC & 100.0 & 100.0 & 0 & 0 & 1.000 & 0.000 \\
\cmidrule(lr){1-8}
\multirow{3}{*}{DeepSeek-R1} & Overall & 98.7 & 98.7 & 1 & 1 & 1.000 & 0.000 \\
& LR & 100.0 & 98.0 & 1 & 0 & 1.000 & $-$0.284 \\
& RC & 96.3 & 100.0 & 0 & 1 & 1.000 & +0.387 \\
\cmidrule(lr){1-8}
\multirow{3}{*}{Kimi K2 Thinking} & Overall & 100.0 & 98.7 & 1 & 0 & 1.000 & $-$0.228 \\
& LR & 100.0 & 98.0 & 1 & 0 & 1.000 & $-$0.284 \\
& RC & 100.0 & 100.0 & 0 & 0 & 1.000 & 0.000 \\
\cmidrule(lr){1-8}
\multirow{3}{*}{QwQ-32B} & Overall & 92.2 & 94.8 & 1 & 3 & .625 & +0.106 \\
& LR & 92.0 & 94.0 & 1 & 2 & 1.000 & +0.079 \\
& RC & 92.6 & 96.3 & 0 & 1 & 1.000 & +0.164 \\
\midrule
\multirow{3}{*}{DS-R1 Distill 7B} & Overall & 55.8 & 49.4 & 17 & 12 & .458 & $-$0.130 \\
& LR & 54.0 & 46.0 & 12 & 8 & .503 & $-$0.160 \\
& RC & 59.3 & 55.6 & 5 & 4 & 1.000 & $-$0.075 \\
\cmidrule(lr){1-8}
\multirow{3}{*}{DS-R1 Distill Llama 8B} & Overall & 53.2 & 64.9 & 6 & 15 & .078 & +0.238 \\
& LR & 46.0 & 58.0 & 4 & 10 & .180 & +0.241 \\
& RC & 66.7 & 77.8 & 2 & 5 & .453 & +0.249 \\
\bottomrule
\end{tabular}
\end{table}
\vspace*{\stretch{1.3}}
\clearpage

\null\vfill
\begin{table}[H]
\caption{Position bias by section on PT 150--159 ($N=1{,}037$; LR $N=633$, RC $N=404$). Original vs.\ shuffled (McNemar's exact test).}
\label{tab:shuffle_section}
\centering
\small
\begin{tabular}{llcccccccl}
\toprule
Model & Section & Original & Shuffled & $b$ & $c$ & $p$ & $h$ & TOST \\
\midrule
\multirow{3}{*}{GPT-5} & Overall & 98.8 & 98.7 & 3 & 2 & 1.000 & $-$0.009 & EQUIV \\
& LR & 99.1 & 98.7 & 2 & 0 & .500 & $-$0.030 & EQUIV \\
& RC & 98.5 & 98.8 & 1 & 2 & 1.000 & +0.021 & EQUIV \\
\cmidrule(lr){1-9}
\multirow{3}{*}{Claude Opus 4} & Overall & 97.9 & 98.0 & 5 & 6 & 1.000 & +0.007 & EQUIV \\
& LR & 97.9 & 97.8 & 3 & 2 & 1.000 & $-$0.011 & EQUIV \\
& RC & 97.8 & 98.3 & 2 & 4 & .688 & +0.036 & NO \\
\cmidrule(lr){1-9}
\multirow{3}{*}{Gemini 2.5 Pro} & Overall & 98.6 & 98.5 & 4 & 3 & 1.000 & $-$0.008 & EQUIV \\
& LR & 99.1 & 98.9 & 2 & 1 & 1.000 & $-$0.016 & EQUIV \\
& RC & 97.8 & 97.8 & 2 & 2 & 1.000 & 0.000 & EQUIV \\
\cmidrule(lr){1-9}
\multirow{3}{*}{DeepSeek-R1} & Overall & 97.7 & 97.4 & 14 & 11 & .690 & $-$0.019 & EQUIV \\
& LR & 97.8 & 97.8 & 7 & 7 & 1.000 & 0.000 & EQUIV \\
& RC & 97.5 & 96.8 & 7 & 4 & .549 & $-$0.045 & NO \\
\cmidrule(lr){1-9}
\multirow{3}{*}{Kimi K2 Thinking} & Overall & 97.2 & 97.0 & 17 & 15 & .860 & $-$0.012 & EQUIV \\
& LR & 97.8 & 96.5 & 14 & 6 & .115 & $-$0.077 & NO \\
& RC & 96.3 & 97.8 & 3 & 9 & .146 & +0.088 & NO \\
\cmidrule(lr){1-9}
\multirow{3}{*}{QwQ-32B} & Overall & 93.2 & 93.2 & 32 & 32 & 1.000 & 0.000 & EQUIV \\
& LR & 92.9 & 92.7 & 20 & 19 & 1.000 & $-$0.006 & NO \\
& RC & 93.8 & 94.1 & 12 & 13 & 1.000 & +0.010 & NO \\
\midrule
\multirow{3}{*}{DS-R1 Distill 7B} & Overall & 56.7 & 57.5 & 150 & 158 & .690 & +0.016 & NO \\
& LR & 54.0 & 56.9 & 81 & 99 & .205 & +0.057 & NO \\
& RC & 60.9 & 58.4 & 69 & 59 & .426 & $-$0.050 & NO \\
\cmidrule(lr){1-9}
\multirow{3}{*}{DS-R1 Distill Llama 8B} & Overall & 63.2 & 63.1 & 130 & 129 & 1.000 & $-$0.002 & NO \\
& LR & 57.0 & 58.3 & 79 & 87 & .587 & +0.026 & NO \\
& RC & 72.8 & 70.5 & 51 & 42 & .407 & $-$0.049 & NO \\
\bottomrule
\end{tabular}
\end{table}
\vfill
\clearpage

\section{Self-Consistency: Per-Section Detailed Results}
\label{app:sc_section}

This appendix expands the self-consistency analysis of Section~\ref{sec:meth_sc}, supplementing Tables~\ref{tab:sc_160}--\ref{tab:sc} with separate Logical Reasoning and Reading Comprehension pass@1 versus SC@5 comparisons, reported on both the Official Test (Table~\ref{tab:sc_section_160}) and PT 150--159 (Table~\ref{tab:sc_section}).

\vfill

\begin{table}[H]
\caption{Self-consistency by section on the Official Test ($N=77$; LR $N=50$, RC $N=27$). McNemar's exact test.}
\label{tab:sc_section_160}
\centering
\small
\begin{tabular}{llccccccc}
\toprule
Model & Section & pass@1 & SC@5 & $\Delta$ & $b$ & $c$ & $p$ & Unanimity \\
\midrule
\multirow{3}{*}{GPT-5} & Overall & 97.4 & 97.4 & 0.0 & 0 & 0 & 1.000 & 97.4\% \\
& LR & 96.0 & 96.0 & 0.0 & 0 & 0 & 1.000 & 98.0\% \\
& RC & 100.0 & 100.0 & 0.0 & 0 & 0 & 1.000 & 96.3\% \\
\cmidrule(lr){1-9}
\multirow{3}{*}{Claude Opus 4} & Overall & 93.5 & 96.1 & +2.6 & 0 & 2 & .500 & 94.8\% \\
& LR & 90.0 & 94.0 & +4.0 & 0 & 2 & .500 & 92.0\% \\
& RC & 100.0 & 100.0 & 0.0 & 0 & 0 & 1.000 & 100.0\% \\
\cmidrule(lr){1-9}
\multirow{3}{*}{Gemini 2.5 Pro} & Overall & 97.4 & 97.4 & 0.0 & 0 & 0 & 1.000 & 97.4\% \\
& LR & 96.0 & 96.0 & 0.0 & 0 & 0 & 1.000 & 96.0\% \\
& RC & 100.0 & 100.0 & 0.0 & 0 & 0 & 1.000 & 100.0\% \\
\cmidrule(lr){1-9}
\multirow{3}{*}{DeepSeek-R1} & Overall & 98.7 & 98.7 & 0.0 & 0 & 0 & 1.000 & 97.4\% \\
& LR & 98.0 & 98.0 & 0.0 & 0 & 0 & 1.000 & 98.0\% \\
& RC & 100.0 & 100.0 & 0.0 & 0 & 0 & 1.000 & 96.3\% \\
\cmidrule(lr){1-9}
\multirow{3}{*}{Kimi K2 Thinking} & Overall & 97.4 & 98.7 & +1.3 & 0 & 1 & 1.000 & 94.8\% \\
& LR & 98.0 & 98.0 & 0.0 & 0 & 0 & 1.000 & 96.0\% \\
& RC & 96.3 & 100.0 & +3.7 & 0 & 1 & 1.000 & 92.6\% \\
\cmidrule(lr){1-9}
\multirow{3}{*}{QwQ-32B} & Overall & 93.5 & 96.1 & +2.6 & 0 & 2 & .500 & 92.2\% \\
& LR & 94.0 & 94.0 & 0.0 & 0 & 0 & 1.000 & 94.0\% \\
& RC & 92.6 & 100.0 & +7.4 & 0 & 2 & .500 & 88.9\% \\
\midrule
\multirow{3}{*}{DS-R1 Distill 7B} & Overall & 51.9 & 55.8 & +3.9 & 7 & 10 & .629 & 22.1\% \\
& LR & 52.0 & 50.0 & $-$2.0 & 5 & 4 & 1.000 & 0.0\% \\
& RC & 51.9 & 66.7 & +14.8 & 2 & 6 & .289 & 0.0\% \\
\cmidrule(lr){1-9}
\multirow{3}{*}{DS-R1 Distill Llama 8B} & Overall & 55.8 & 59.7 & +3.9 & 3 & 6 & .508 & 37.7\% \\
& LR & 50.0 & 54.0 & +4.0 & 3 & 5 & .727 & 0.0\% \\
& RC & 66.7 & 70.4 & +3.7 & 0 & 1 & 1.000 & 0.0\% \\
\bottomrule
\end{tabular}
\end{table}
\vspace*{\stretch{1.3}}
\clearpage

\null\vfill
\begin{table}[H]
\caption{Self-consistency by section on PT 150--159 ($N=1{,}037$; LR $N=633$, RC $N=404$). McNemar's exact test.}
\label{tab:sc_section}
\centering
\small
\begin{tabular}{llcccccccc}
\toprule
Model & Section & pass@1 & SC@5 & $\Delta$ & $b$ & $c$ & $p$ & $h$ & Unanimity \\
\midrule
\multirow{3}{*}{GPT-5} & Overall & 97.3 & 99.0 & +1.7 & 0 & 18 & $<$.001 & +0.133 & 92.8\% \\
& LR & 97.5 & 99.2 & +1.7 & 0 & 11 & .001 & +0.141 & 93.4\% \\
& RC & 97.0 & 98.8 & +1.7 & 0 & 7 & .016 & +0.123 & 91.8\% \\
\cmidrule(lr){1-10}
\multirow{3}{*}{Claude Opus 4} & Overall & 97.1 & 98.0 & +0.9 & 4 & 13 & .049 & +0.056 & 91.7\% \\
& LR & 97.3 & 98.3 & +0.9 & 2 & 8 & .109 & +0.065 & 91.9\% \\
& RC & 96.8 & 97.5 & +0.7 & 2 & 5 & .453 & +0.045 & 91.3\% \\
\cmidrule(lr){1-10}
\multirow{3}{*}{Gemini 2.5 Pro} & Overall & 97.6 & 98.6 & +1.1 & 3 & 14 & .013 & +0.079 & 93.7\% \\
& LR & 98.4 & 99.1 & +0.6 & 2 & 6 & .289 & +0.057 & 94.3\% \\
& RC & 96.3 & 98.0 & +1.7 & 1 & 8 & .039 & +0.105 & 92.8\% \\
\cmidrule(lr){1-10}
\multirow{3}{*}{DeepSeek-R1} & Overall & 97.1 & 97.8 & +0.7 & 2 & 9 & .065 & +0.043 & 94.9\% \\
& LR & 97.5 & 98.1 & +0.6 & 1 & 5 & .219 & +0.043 & 94.8\% \\
& RC & 96.5 & 97.3 & +0.7 & 1 & 4 & .375 & +0.043 & 95.0\% \\
\cmidrule(lr){1-10}
\multirow{3}{*}{Kimi K2 Thinking} & Overall & 97.2 & 98.4 & +1.2 & 2 & 14 & .004 & +0.079 & 92.8\% \\
& LR & 97.5 & 98.3 & +0.8 & 2 & 7 & .180 & +0.055 & 92.7\% \\
& RC & 96.8 & 98.5 & +1.7 & 0 & 7 & .016 & +0.116 & 92.8\% \\
\cmidrule(lr){1-10}
\multirow{3}{*}{QwQ-32B} & Overall & 94.5 & 94.9 & +0.4 & 7 & 11 & .481 & +0.017 & 88.9\% \\
& LR & 94.2 & 94.2 & 0.0 & 5 & 5 & 1.000 & 0.000 & 88.0\% \\
& RC & 95.0 & 96.0 & +1.0 & 2 & 6 & .289 & +0.048 & 90.3\% \\
\midrule
\multirow{3}{*}{DS-R1 Distill 7B} & Overall & 58.7 & 62.6 & +3.9 & 58 & 98 & .002 & +0.079 & 35.3\% \\
& LR & 57.5 & 61.8 & +4.3 & 32 & 59 & .006 & +0.087 & 32.9\% \\
& RC & 60.6 & 63.9 & +3.2 & 26 & 39 & .136 & +0.066 & 39.1\% \\
\cmidrule(lr){1-10}
\multirow{3}{*}{DS-R1 Distill Llama 8B} & Overall & 63.9 & 68.9 & +4.9 & 48 & 99 & $<$.001 & +0.104 & 44.6\% \\
& LR & 58.9 & 64.5 & +5.5 & 33 & 68 & .001 & +0.114 & 40.4\% \\
& RC & 71.8 & 75.7 & +4.0 & 15 & 31 & .026 & +0.090 & 51.0\% \\
\bottomrule
\end{tabular}
\end{table}
\vfill
\clearpage


\end{document}